\documentclass[a4paper, 10pt, twocolumn, journal, final]{ieeeconf}
\IEEEoverridecommandlockouts  
\usepackage[utf8]{inputenc}
\usepackage{graphicx} 
\usepackage{silence}
\usepackage{caption}
\usepackage{epsfig} 
\usepackage{rotating}
\usepackage{fontawesome}  
\usepackage{mathptmx} 
\usepackage{times}  
\usepackage{amsmath} 
\usepackage{amssymb} 
\usepackage{dsfont} 
\usepackage{mathtools}
\usepackage{empheq}
\usepackage{eucal}
\usepackage{bm} 
\usepackage{pgfgantt}
\usepackage{booktabs}  
\usepackage[flushleft]{threeparttable}
\usepackage[colorlinks=true, citecolor = blue, urlcolor = blue, linkcolor = blue]{hyperref}

\newcommand{\vecgreek}[1]{\pmb{#1}}

\usepackage{letterbib}
\usepackage{scalerel}
\usepackage[authoryear, round]{natbib}
\definecolor{barblue}{RGB}{153,204,254}
\definecolor{groupblue}{RGB}{51,102,254}
\definecolor{linkred}{RGB}{165,0,33}

\renewcommand\labelenumi{(\roman{enumi})}
\renewcommand\theenumi\labelenumi 

\usepackage{lmodern} %Font
\usepackage[nolists, tablesfirst]{endfloat}
\DeclareDelayedFloatFlavor{sidewaystable}{table}
\DeclareDelayedFloatFlavor{sidewaysfigure}{figure}

\renewcommand{\thefigure}{\arabic{figure}}
\renewcommand{\thetable}{\arabic{table}}
\usepackage{subcaption}

\title{\LARGE{\bf{Distributional Gradient Boosting Machines}}

\vspace{1em}

}
\author{\parbox{3 in}{\centering Alexander März$^{\aleph_{\scaleto{0\mathstrut}{2pt}}}$
        \thanks{$^{\aleph_{\scaleto{0\mathstrut}{4pt}}}$ \hspace{-0.8em} Author for correspondence: \href{mailto:alex.maerz@gmx.net}{alex.maerz@gmx.net}}\\
        {\small \textcolor{white}{Georg-August-Universit{\"a}t G{\"o}ttingen}} \\
    }      
        \hspace*{0.5in}        
        \parbox{3 in}{\centering Thomas Kneib \\
        {\small Georg-August-Universit{\"a}t G{\"o}ttingen} \\
    }
}

\begin{document}
\maketitle
\thispagestyle{empty}
%\pagestyle{empty}

%%%%%%%%%%%%%%%%%%%%%%%%%%%%%%%%%%%%%%%%%%%%%%%%%%%%%%%%%%%%%%%%%%%%%%%%%%%%%%%%
\begin{abstract}
We present a unified probabilistic gradient boosting framework for regression tasks that models and predicts the entire conditional distribution of a univariate response variable as a function of covariates. Our likelihood-based approach allows us to either model all conditional moments of a parametric distribution, or to approximate the conditional cumulative distribution function via Normalizing Flows. As underlying computational backbones, our framework is based on XGBoost and LightGBM. Modelling and predicting the entire conditional distribution greatly enhances existing tree-based gradient boosting implementations, as it allows to create probabilistic forecasts from which prediction intervals and quantiles of interest can be derived. Empirical results show that our framework achieves state-of-the-art forecast accuracy. 
\end{abstract}

%%%%%%%%%%%%%%%%%%%%%%%%%%%%%%%%%%%%%%%%%%%%%%%%%%%%%%%%%%%%%%%%%%%%%%%%%%%%%%%%
\vspace{1em}

%%%%%%%%%%%%%%%%%%%%%%%%%%%%%%%%%%%%%%%%%%%%%%%%%%%%%%%%%%%%%%%%%%%%%%%%%%%%%%%%
\section{Introduction}

The development of modelling approaches that approximate and describe the data generating processes underlying the observed data in as much detail as possible is a guiding principle in both statistics and machine learning. We therefore strongly agree with the statement of \cite{Hothorn.2014} that \textit{'the ultimate goal of any regression analysis is to obtain information about the entire conditional distribution $F_{Y}(y|\mathbf{x})$ of a response given a set of explanatory variables'}.\footnote{We denote $\mathbb{P}(Y \leq y | \mathbf{X} = \mathbf{x}) = F_{Y}(y|\mathbf{x})$ the conditional cumulative distribution function of a potentially continuous, discrete or mixed discrete-continuous response $Y$ given explanatory variables $\mathbf{X} = \mathbf{x}$. $f_{Y}(y|\mathbf{x})$ denotes the conditional density of $Y$. The conditional quantile function is denoted as $F^{-1}_{Y}(y|\mathbf{x})$.} It has not been too long, though, that most regression models focused on estimating the conditional mean $\mathbb{E}(Y|\mathbf{X} = \mathbf{x})$ only, implicitly treating higher moments of the conditional distribution $F_{Y}(y|\mathbf{x})$ as fixed nuisance parameters. As such, models that minimize an $\ell_{2}$-loss for the conditional mean are not able to fully exploit the information contained in the data, since this is equivalent of assuming a Normal distribution with constant variance. In real world situations, however, the data generating process is usually less well behaved, exhibiting characteristics such as heteroskedasticity or varying degrees of skewness and kurtosis. As an example, the data of the M5 forecasting competition can be characterized by a high degree of overdispersion, as well as intermittent and sporadic demand behaviour \citep{Ziel.2021}. In recent years, however, there has been a clear shift in both academic and corporate research towards modelling the entire conditional distribution. This change in attention is most evident in the M5 forecasting competition, which differed from previous ones in the sense that it consisted of two parallel competitions: in addition to providing accurate point forecasts, participants were also asked to forecast nine different quantiles to approximate the distribution of future sales.\footnote{For details on the M5 competition, see \cite{Makridakis.2021a} and \cite{Makridakis.2021}.} 

Initiated by the seminal paper of \cite{Salinas.2020}, recent advances in probabilistic time series forecasting have been predominantly presented in the context of deep learning, where forecasting has greatly benefited from a global modelling approach, with the parameters of the model being learned across a set of (related) time series, instead of training a model for each time series individually.\footnote{For an overview of neural operational and strategic time series forecasting approaches, see \citet{Januschowski.2021, Januschowski.2020, Januschowski.2019} and the references therein.} Compared with the vast body of literature on neural probabilistic time series forecasting, tree-based models have received comparatively little attention. Yet, the M5 competition has demonstrated tree-based models to be very competitive for operational forecasting tasks, making them a viable alternative to deep learning approaches. \cite{Januschowski.2021} provide an overview of the advantages of tree-based models to explain why they ranked so high for both tracks of the M5 competition. Among the most salient features of tree-based models, the authors include their robustness and their comparatively low sensitivity towards hyper-parameters, handling of sparse and intermittent targets, built-in handling of missing and categorical features, their interpretability, and the fact that both the inputs and the target variable do not have to be scaled. All this makes tree-based models a ready to use and out-of-the box competitive class of models. 

Besides these advantages, current implementations of winning tree-based models, such as XGBoost of \cite{Chen.2016} and LightGBM introduced by \cite{Ke.2017}, do not readily provide probabilistic forecasts. With this paper, we respond to the need for turning tree-based point forecasts into probabilistic ones and present a unified tree-based probabilistic gradient boosting framework for regression tasks. Using XGBoost and the M5-forecasting competition winning LightGBM model as computational backbones, our approach allows us to either model all conditional moments of a parametric distribution, or to approximate the conditional cumulative distribution function (CDF) via a novel Normalizing Flow based approach. To the best of our knowledge, we are among the first to use Normalizing Flows in a gradient boosting framework. Unlike other methods, our framework is entirely likelihood based, where both sampling and evaluation are efficient and exact using tractable likelihood functions. 

The remainder of this paper is organized as follows: Section \ref{sec:dgbm} introduces our Distributional Gradient Boosting Machine (DGBM) framework and Section \ref{sec:literature} presents an overview of related research branches. Section \ref{sec:applications} presents both a simulation study and real world examples that demonstrate the effectiveness of our framework. Section \ref{sec:conclusion} concludes.

%%%%%%%%%%%%%%%%%%%%%%%%%%%%%%%%%%%%%%%%%%%%%%%%%%%%%%%%%%%%%%%%%%%%%%%%%%%%%%%%%
\section{DGBM: Distributional Gradient Boosting Machines} \label{sec:dgbm}

Assessing the uncertainty attached to an outcome is essential for any decision making. This is especially true for forecasting tasks, where modelling the full distribution allows to generate different trajectories of potential future states. While the notion of uncertainty is ubiquitous, it is usually not well defined. To allow for a more nuanced view on uncertainty, we follow \cite{Hullermeier.2021} and distinguish between two sources of uncertainty for any machine learning task: while epistemic uncertainty accounts for uncertainty in the model that can generally be reduced given sufficient data, this paper is concerned with aleatoric uncertainty, which reflects the randomness inherent in observations. This type of uncertainty can be captured by modelling a conditional outcome probability distribution that is dependent on features \citep{Baumann.2021}. 

While there are several ways to arrive at probabilistic forecasts, modelling conditional quantiles as introduced by \cite{Koenker.1978} is the most common choice, as they are both easy to interpret and provide insights into different parts of the response distribution. One common way, also applied by some contestants during the M5 competition, is to estimate one model for each quantile separately. While this approach is easy to scale and parallelize, it has the disadvantage of potential quantile crossing, especially when a dense set of quantiles is to be modelled and forecasted. Even though a penalty term can be added during model training, it is not guaranteed that quantiles are non-crossing for out-of-sample forecasts. Moreover, maintaining multiple models for each quantile can very easily become prohibitive, especially when it comes to variable selection and hyper-parameter tuning.

To avoid any of these problems, we suggest to directly model the entire conditional distribution of a response with a single model and to derive quantiles from the forecasted distribution. In this section, we present two alternatives of how to estimate the response distribution using gradient boosted decision trees, either via: 

\begin{enumerate}
	\item assuming a parametric form of the response distribution, modelling all distributional parameters (Section \ref{sec:gbmlss})
	\item approximating the conditional cumulative distribution function via Normalizing Flows (Section \ref{sec:nfboost})
\end{enumerate}

\noindent The strict monotonicity of the forecasted distribution guarantees that for well-calibrated models quantiles are in fact non-crossing. For each of the two alternatives, our framework uses XGBoost (XGB) and LightGBM (LGB) as underlying computational engines. 

\subsection{GBMLSS: Gradient Boosting Machines for Location, Scale and Shape} \label{sec:gbmlss}

Probabilistic forecasts are predictions in the form of a probability distribution, rather than a single point estimate only. In this context, the introduction of Generalized Additive Models for Location Scale and Shape (GAMLSS) by \citet{Rigby.2005} has stimulated a lot of research and culminated in a new research branch that focuses on modelling the entire conditional distribution in dependence of covariates. This section introduces the general idea of distributional modelling.\footnote{For a more in-depth introduction, we draw the reader's attention to \citet{Rigby.2005, Klein.2015c, Klein.2015b, Stasinopoulos.2017}.}

In its original formulation, GAMLSS assume a univariate response to follow a distribution $\mathcal{D}$ that depends on up to four parameters, i.e., $y_{i} \stackrel{ind}{\sim} \mathcal{D}(\mu_{i}, \sigma^{2}_{i}, \nu_{i}, \tau_{i}), i=1,\ldots,n$, where $\mu_{i}$ and $\sigma^{2}_{i}$ are often location and scale parameters, respectively, while $\nu_{i}$ and $\tau_{i}$ correspond to shape parameters such as skewness and kurtosis. Hence, the framework allows to model not only the mean (or location) but all parameters as functions of explanatory variables.\footnote{It is important to note that distributional modelling implies that observations are independent, but not necessarily identical realizations $y \stackrel{ind}{\sim} \mathcal{D}\big(\mathbf{\theta}(\mathbf{x})\big)$, since all distributional parameters $\mathbf{\theta}(\mathbf{x})$ are related to and allowed to change with covariates.} In contrast to Generalized Linear (GLM) and Generalized Additive Models (GAM), the assumption of the response distribution belonging to an exponential family is relaxed in GAMLSS and replaced by a more general class of distributions, including highly skewed and/or kurtotic continuous, discrete and mixed discrete, as well as zero-inflated distributions.\footnote{While the original formulation of GAMLSS in \citet{Rigby.2005} suggests that any distribution can be described by location, scale and shape parameters, it is not necessarily true that the observed data distribution can actually be characterized by all of these parameters. Hence, we follow \citet{Klein.2015b} and use the term distributional modelling and GAMLSS interchangeably.} 

From a frequentist point of view, distributional modelling can be formulated as follows

\begin{empheq}[left=y_{i} \stackrel{ind}{\sim} \mathcal{D} \empheqbiglparen, right=\empheqbigrparen]{align}
	h_{1}(\theta_{i1}) &= \eta_{i1} \nonumber\\ 
	h_{2}(\theta_{i2}) &= \eta_{i2}  \label{eq:dist_model}  \\ 
	\vdots \nonumber \\                        
	h_{K}(\theta_{iK}) &= \eta_{iK} \nonumber 
\end{empheq}

\noindent for $i = 1, \ldots, n$, where $\mathcal{D}$ denotes a parametric distribution for the response $\textbf{y} = (y_{1}, \ldots, y_{n})^{\prime}$ that depends on $K$ distributional parameters $\theta_{k}$, $k = 1, \ldots, K$, and with $h_{k}(\cdot)$ denoting a known function relating distributional parameters to predictors $\vecgreek{\eta}_{k}$. In its most generic form, the predictor $\vecgreek{\eta}_{k}$ is given by

\begin{equation}
\vecgreek{\eta}_{k} = f_{k}(\mathbf{x}), \qquad k = 1, \ldots, K \label{eq:gamlss}
\end{equation} 

\noindent Within the original distributional regression framework, the functions $f_{k}(\cdot)$ usually represent a combination of linear and GAM-type predictors, which allows to estimate linear effects or categorical variables, as well as highly non-linear and spatial effects using a Spline-based basis function approach.\footnote{See \citealt{Fahrmeir.2011} and \citealt{Fahrmeir.2013} for further details.} Concerning the estimation of distributional regression, it relies on the availability of first and second order derivatives of the log-likelihood function needed for Fisher-scoring type algorithms. 

The predictor specification in Equation \eqref{eq:gamlss} is generic enough to use tree-based models as well, which allows us to extend  XGBoost and LightGBM to a probabilistic framework. We term our approach Gradient Boosting Machines for Location, Scale and Shape (GBMLSS) and interpret the loss function from a statistical perspective by formulating empirical risk minimization as Maximum Likelihood estimation. As outlined in \citet{Maerz.2019, Maerz.2020}, GBMLSS require the specification of a suitable distribution from which gradients and hessians are derived.\footnote{It is important to note that both XGBoost and LightGBM approximate the loss function with a second-order expansion, which requires gradients and hessians to be non-zero and defined everywhere. The fact that the quantile loss has no well-defined second order derivative makes quantile regression using the pinball loss not feasible using Newton boosting without a workaround. Some implementations approximate the hessian with 1, which collapses Newton boosting to ordinary gradient descent boosting. However, \citet{Sigrist.2021} provides empirical evidence that Newton Boosting generally outperforms gradient boosting on the majority of data sets used for the comparison. \citet{Sigrist.2021} mainly attributes the advantage of Newton over gradient boosting to the variability in the hessians, i.e., the more variation there is in the second order terms, the more pronounced is the difference between the two approaches and the more likely is Newton to outperform gradient boosting.} These represent the partial first and second order derivatives of the log-likelihood with respect to the distributional parameter $\theta_{k}$ of interest. GBMLSS are based on multi-parameter optimization, where a separate tree is grown for each of the $k = 1, \ldots, K$ distributional parameters. Estimation of gradients and hessians, as well as the evaluation of the loss function is done simultaneously for all distributional parameters.\footnote{For training GBMLSS, we leverage the link to multiclass-classification where, similar to our approach, a separate tree is grown for each class, and where the cross-entropy loss is used as an evaluation criteria for the different class-probabilities.} To improve on the convergence and stability of GBMLSS estimation, unconditional Maximum Likelihood estimates of the parameters $\theta_{k}$, $k = 1, \ldots, K$ are used as offset values. In addition to exact gradients and hessians, GBMLSS also support automatic differentiation of any twice-differentiable loss-function; a property that we leverage for estimating Normalizing Flow based models as introduced in the next section.\footnote{In its current implementation, our DGBM framework supports automatic differentiation using PyTorch \citep{Paszke.2019} and TensorFlow \citep{Abadi.2015}. The flexibility offered by automatic differentiation allows one to implement novel parametric distributions for which gradients and hessians are difficult to derive, or to add additional constraints to the loss function. As an example, since our framework also allows the estimation of several expectiles simultaneously, one can add a penalty to avoid crossing of expectiles.} Using the Shapley-Value approach of \citet{Lundberg.2020, Lundberg.2017}, GBMLSS offer the additional advantage of providing attribute importance and partial dependence plots for all distributional parameters individually.

\subsection{NFBoost: Normalizing Flow Boosting} \label{sec:nfboost}
The previous section has shown that the GBMLSS framework provides a high level of flexibility due to its ability to model a variety 
of complex distributions. However, there might also be situations in which a parametric distribution might not be flexible enough to provide a reasonable approximation to the data at hand. For such cases, it may be preferable to relax the assumption of a parametric distribution and approximate the data non-parametrically. 

While there are several ways for estimating the conditional cumulative distribution function, we propose to use conditional Normalizing Flows (NF) for their ability to fit complex and high dimensional distributions with only a few parameters.\footnote{See \cite{Papamakarios.2021} and \cite{Kobyzev.2020} for a more detailed overview.} The principle that underlies Normalizing Flows is to turn a simple base distribution, e.g., $F_{Z}(\mathbf{z}) = N(0,1)$, into a more complex and realistic distribution of the target variable $F_{Y}(\mathbf{y})$ by applying several bijective transformations $h_{j}$, $j = 1, \ldots, J$ to the variable of the base distribution \citep{Ruegamer.2022}

\begin{equation}
	\mathbf{y} = h_{J} \circ h_{J-1} \circ \cdots \circ h_{1}(\mathbf{z})
\end{equation}

\noindent Based on the complete transformation function $h=h_{J}\circ\ldots\circ h_{1}$, the density of $\mathbf{y}$ is then given by the change of variables theorem

\begin{equation}
	f_{Y}(\mathbf{y}) = f_{Z}\big(h(\mathbf{y})\big) \cdot \Bigg|\frac{\partial h(\mathbf{y})}{\partial \mathbf{y}}\Bigg| \label{eq:change_variable_theorem}
\end{equation}

%\begin{equation}
%	f_{Y}(\mathbf{y}) = f_{Z}(h(\mathbf{y}))|\mbox{det}\nabla h(\mathbf{y})|^{-1} \label{eq:change_variable_theorem}
%\end{equation}

\noindent where scaling with the Jacobian determinant $|h^{\prime}(\mathbf{y})| = |\partial h(\mathbf{y}) / \partial \mathbf{y}|$ ensures $f_{Y}(\mathbf{y})$ to be a proper density integrating to one.

Our Normalizing Flow Boosting approach (NFBoost) is based on Bernstein-Polynomial Normalizing Flows introduced by \cite{Sick.2021} and further extended by \citet{Duerr.2022, Arpogaus.2021}. This type of Normalizing Flow is in turn based on Conditional Transformation Models (CTMs) originally introduced by \cite{Hothorn.2014}.\footnote{\cite{Klein.2020} and \cite{Sick.2021} highlight the close resemblance between Normalizing Flows and Conditional Transformation Models, even though and in contrast to the initial formulation of Conditional Transformation Models, Normalizing Flows usually consist of several chained transformations.} As for Normalizing Flows, Conditional Transformation Models transform a simple base distribution $F_{Z}$ into a more complex and realistic target distribution $F_{Y}$ via a monotonic conditional transformation function $h(y|\mathbf{x})$

\begin{equation}
    F_{Y|\mathbf{x}}(y) = \mathbb{P}(Y \leq y |\mathbf{x}) = F_{Z}(h(y|\mathbf{x}))
\end{equation}

\noindent that is learnt from the data \citep{Baumann.2021}. However, instead of a transformation from $\mathbf{z}$ to $\mathbf{y}$, Transformation Models define an inverse flow $h(\mathbf{y}) = \mathbf{z}$ \citep{Ruegamer.2022}. For continuous $Y$, the transformation function is typically approximated using Bernstein-Polynomials with covariate dependent basis coefficients $\vartheta_{m}(\mathbf{x})$

\begin{equation}
	h_{\vartheta}(\tilde{y}|\mathbf{x}) = \frac{1}{M+1}\sum^{M}_{m=0}\mbox{Be}_{m}(\tilde{y})\vartheta_{m}(\mathbf{x}) \label{eq:BP}
\end{equation}

\noindent with $\mbox{Be}_{m}(\tilde{y}) = f_{m+1,M-m+1}(\tilde{y})$ being $M+1$ Beta-densities and $\tilde{y}$ being a re-scaled version of the original target $y$ to ensure $\tilde{y} \in [0,1]$ necessary for the Beta-densities. $\vartheta_{0} < \vartheta_{1} < \ldots < \vartheta_{M}$ ensures monotonicity of $F_{Z}(h(y|\mathbf{x}))$ and hence of the estimated CDF. Their properties to uniformly approximate any function in $[0,1]$, as well as their computational efficiency make Bernstein-Polynomials a reasonable choice \citep{Sick.2021}. 

Even though Bernstein-Polynomials are already very flexible to approximate any CDF, \citet{Duerr.2022, Arpogaus.2021, Sick.2021} add two additional scale and shift transformations before and after the Bernstein-Polynomial transformation. This results in more efficient training and allows to more easily choose a simple base distribution $F_{Z}$ that has support outside $[0,1]$, such as the Standard-Normal \citep{Duerr.2022}. The first transformation $\sigma \circ f_{1}: \tilde{y} = \sigma \big(a_{1}(\mathbf{x}) \cdot y - b_{1}(\mathbf{x}) \big)$ \noindent scales and shifts the original response $y$ and transforms it to $[0,1]$ via the sigmoid function $\sigma$. The second transformation $f_{BP}: \tilde{z} = h_{\vartheta(\mathbf{x})}(\tilde{y})$ is the one defined in Equation \eqref{eq:BP} using the transformed target $\tilde{y}$ from $f_{1}$. The final transformation $f_{3}: z = a_{2}(\mathbf{x}) \cdot \tilde{z} - b_{2}(\mathbf{x})$ is again a scale and shift transformation into the range of the Standard-Normal. The total set of transformations $h_{\xi(\mathbf{x})}(y): y \rightarrow z$ is given by chaining all three together \citep{Sick.2021} 

\begin{align}
	z &= h_{\xi(\mathbf{x})}(y) \label{eq:bf_flow}   \\ 
	&= {f_{3}}_{(a_{2},b_{2})} \circ {f_{BP}}_{(\vartheta_{0}, \ldots, \vartheta_{M})} \circ \sigma \circ {f_{1}}_{(a_{1}, b_{1})}(y) \nonumber	
%		&= {f_{3}}_{(a_{2}(\mathbf{x}),b_{2}(\mathbf{x}))} \circ {f_{BP}}_{(\vartheta_{0}(\mathbf{x}), \ldots, \vartheta_{M}(\mathbf{x}))} \circ \sigma \circ {f_{1}}_{(a_{1}(\mathbf{x}), b_{1}(\mathbf{x}))}(y) \nonumber	
\end{align}

\noindent All parameters in $\xi(\mathbf{x}) = \{a_{1}(\mathbf{x}), b_{1}(\mathbf{x}), \vartheta_{1}(\mathbf{x}), \ldots, \vartheta_{M}(\mathbf{x}), a_{2}(\mathbf{x}), b_{2}(\mathbf{x})\}$ are functions of covariates. Based on the change of variables theorem in Equation \eqref{eq:change_variable_theorem}, parameters in $\xi(\mathbf{x})$ are estimated using the following likelihood function 

\begin{equation}
f_{Y}(y|\mathbf{x}) = f_{Z}\big(h_{\xi(\mathbf{x})}(y)\big) \cdot \Bigg|\frac{\partial h_{\xi(\mathbf{x})}(y)}{\partial y}\Bigg|
\end{equation}

\noindent To ensure the chain of transformation in Equation \eqref{eq:bf_flow} to be invertible, all individual components need to be strictly monotonous \citep{Arpogaus.2021}.\footnote{However, a closed-form solution for the inversion of higher-order Bernstein polynomials is not known so that, based on \cite{Farouki.2012}, \cite{Arpogaus.2021} use cubic B-Splines to approximate its inverse.} To ensure monotonicity for $f_{1}$ and $f_{3}$, $a_{1}$ and $a_{2}$ need to be positive. For the $f_{BP}$ transformation, the parameters $\vartheta_{m}$ need to be increasing. To meet these constraints, \citet{Duerr.2022, Arpogaus.2021, Sick.2021} suggest applying the \textsl{softplus} functions to the scale parameters, i.e., $a^{\prime}_{l} = \textsl{softplus}(a_{l})$ for $l = \{1,2\}$ and $\vartheta^{\prime}_{m} = \vartheta^{\prime}_{m-1} + \textsl{softplus}(\vartheta_{m})$ for $m=1,\ldots,M$ and $\vartheta^{\prime}_{0} = \vartheta_{0}$ to ensure the basis coefficients of the Bernstein-Polynomial transformation to be increasing.\footnote{Other functions such as \textsl{softmax} or \textsl{exp} are also viable options for transforming all parameters in $\xi(\mathbf{x})$.} As for GBMLSS, we use unconditional Maximum Likelihood estimates of the parameters $\xi(\mathbf{x})$ as offset values to improve on the convergence and stability of NFBoost.\footnote{For NFBoost, we use the L-BFGS algorithm to estimate unconditional parameters $\xi$ as offset parameters. To set starting values for the L-BFGS algorithm, we have experimented with several initialization methods such as Xavier \citep{Glorot.2010} or He-initialization of \citep{He.2015}, and found that a simple Standard-Normal initialization leads to the best results.} For the implementation of NFBoost, we resort to the TensorFlow Probability Bernstein-Polynomial Normalizing Flow version of \citet{Arpogaus.2022, Arpogaus.2021}.\footnote{The code is available at \url{https://github.com/MArpogaus/TensorFlow-Probability-Bernstein-Polynomial-Bijector}.}

Similar to GBMLSS introduced in the previous section, NFBoost is based on XGBoost and LightGBM as computational backbones for estimating $\xi(\mathbf{x})$. As such, our DGBM framework offers XGBoost's and LightGBM's full functionality.\footnote{We have implemented our  framework in such a way that XGBoost and LightGBM remain largely unchanged, i.e., DGBM models are wrappers around the initial implementations. The only requirement for using GPU and distributed versions, e.g., via Dask or Ray, is that they need to support custom evaluation metric and objective functions.} The code of DGBM will be made available on our Git-repo \faGithub\href{https://github.com/StatMixedML/DGBM}{https://github.com/StatMixedML/DGBM} at the time of the final publication of the paper.

%, e.g., monotonic or feature interaction constraints, estimating random forest type models, as well as their efficiency for large-scale datasets

%%%%%%%%%%%%%%%%%%%%%%%%%%%%%%%%%%%%%%%%%%%%%%%%%%%%%%%%%%%%%%%%%%%%%%%%%%%%%%%%%
\section{Related Research} \label{sec:literature}

While neural probabilistic forecasting models and its literature have rapidly advanced (see \cite{Januschowski.2021} or \cite{Rasul.2021} for an overview), this section focuses on tree-based approaches for probabilistic and distributional modelling. Exceptions are the neural network based Bernstein Flows of \citet{Duerr.2022,  Arpogaus.2021,  Sick.2021}, on which our approach is based. The authors have successfully applied neural Bernstein Flow models for short-term load forecasting, UCI datasets, as well as for approximating posteriors in Variational Bayes inference. Furthermore, \citet{Ruegamer.2022, Baumann.2021} show state-of-the-art performance of Conditional Transformation based neural networks applied to time series, as well as UCI datasets.

Turning to tree-based models, \cite{Duan.2020} introduce an approach for probabilistic gradient boosting using natural gradients. Similar to GBMLSS, the authors assume a parametric form of the distribution to estimate conditional distributional parameters. By treating leaf weights in each tree as random variables, \cite{Sprangers.2021} estimate mean and variance parameters via stochastic tree ensemble update equations. Using these learned moments allows the authors to sample from a specified distribution after training and to model complex distributions. Since the approach does not rely on multi-parameter boosting, it is computationally efficient and scales well even for large datasets. However, the framework of \cite{Sprangers.2021} is based upon minimizing the mean-squared error (MSE) and is restricted to distributions that can be parametrized using location and scale parameters only. 

In a very recent paper, \citet{Hasson.2021, Januschowski.2021} develop an approach that allows tree-based approaches to be transformed into probabilistic models. By grouping training data whose predictions are sufficiently close, the authors use the resulting bins of true values in the training set as the predicted distributions. Applying their model to the bottom level time series of the M5-competition data, \cite{Januschowski.2021} show that their model ranks among the top 5 in the uncertainty competition. By embedding tree-based regression models into a well-defined theory of conditional inference procedures, where significance tests are used for recursive partitioning, \cite{Schlosser.2018} estimate the parameters of a distribution using Random Forests, whereas \cite{Hothorn.2021} use conditional inference trees and forests for estimating Conditional Transformation models. Based on a statistical view on boosting, \cite{Hothorn.2019} estimates CTMs using component-wise gradient boosting. 

Other approaches include Quantile Regression Forests introduced by \cite{Meinshausen.2006} and the Generalized Regression Forests of \cite{Athey.2019} that use a local nearest neighbour weights approach to estimate different points of the conditional distribution. Bayesian Additive Regression Trees (BART) of \cite{Chipman.2010} are another very interesting strand of literature, as they take a Bayesian view of estimating decision trees and forests. To accommodate for heteroskedastic settings, \cite{Pratola.2020} recently introduced a
heteroscedastic version of BART. In a recent paper, \cite{Giaquinto.2020} combine gradient boosting with Normalizing Flows and apply it to density estimation, as well as to generative modelling of images in combination with a Variational Autoencoder. \cite{Friedman.2020} introduces a boosting framework using contrast trees to estimate the full conditional probability distribution without any assumptions regarding its shape, form, or parametric representation.

%%%%%%%%%%%%%%%%%%%%%%%%%%%%%%%%%%%%%%%%%%%%%%%%%%%%%%%%%%%%%%%%%%%%%%%%%%%%%%%%%
\section{Applications} \label{sec:applications}

In this section, we present both a simulation study and real-world examples that demonstrate the functionality of our framework.

\subsection{Simulation Study} \label{sec:simulation} 

\begin{figure}[h!]
	\centering
	\includegraphics[width=0.7\linewidth]{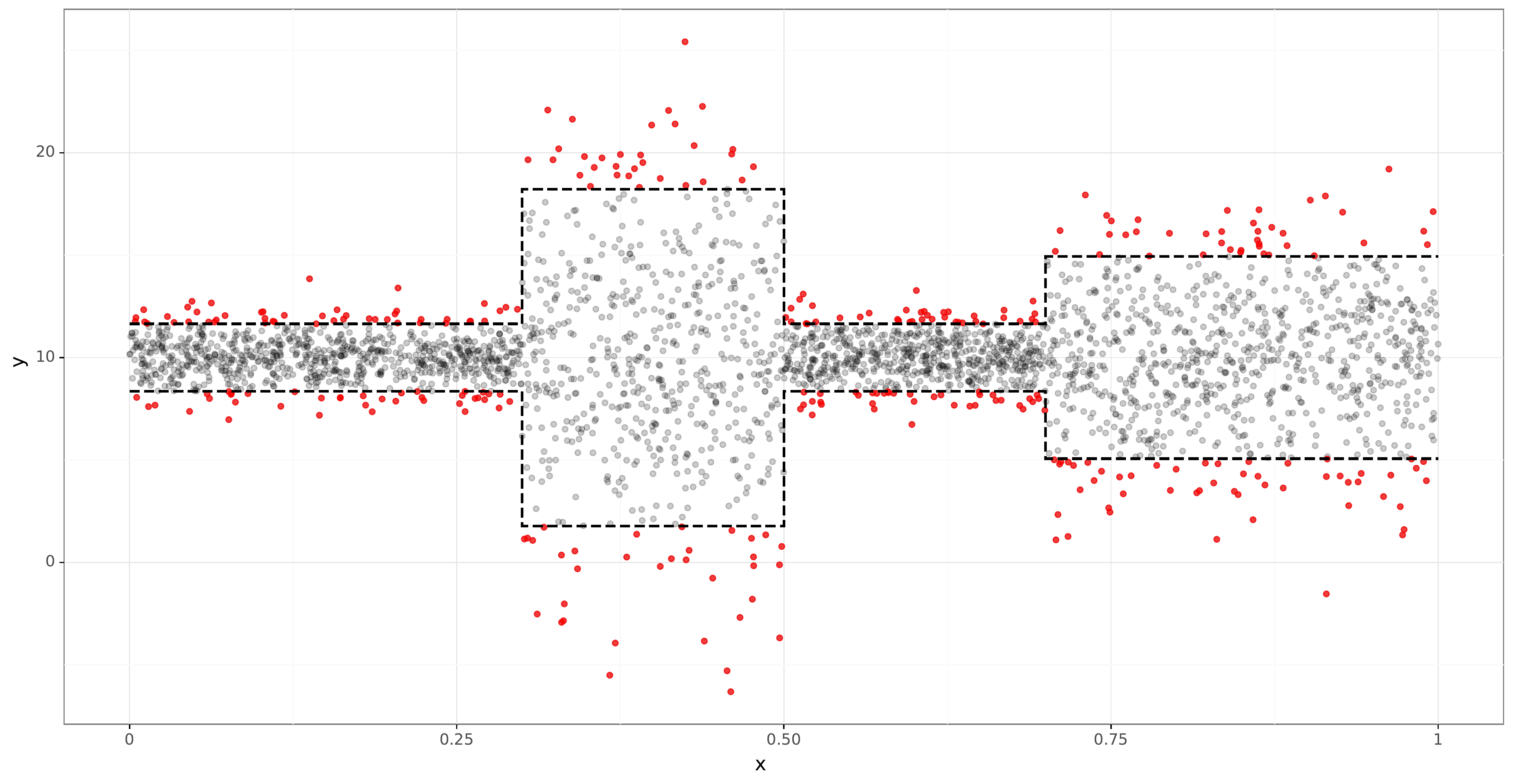}
	\caption{Simulated Train Dataset with 7,000 observations $y \sim \mathcal{N}(10,(1 + 4(0.3 < x < 0.5) + 2(x > 0.7))$. Points outside the 5\% and 95\% quantile are coloured in red. The black dashed lines depict the actual 5\% and 95\% quantiles. Besides the only informative predictor $x$, we have added $X_{1}, \ldots, X_{10}$ as noise variables.}
	\label{fig:sim_data}
\end{figure}

We start with a simulated data set presented in Figure \ref{fig:sim_data} that exhibits a considerable amount of heteroskedasticity, where the interest lies in predicting the 5\% and 95\% quantiles.\footnote{For the simulation, we slightly modify the example presented in \citet{Hothorn.2021}.} The dots in red show points that lie outside the 5\% and 95\% quantiles, which are indicated by the black dashed lines.

\noindent As splitting procedures, that are internally used to construct trees, can detect changes in the mean only, standard tree-based implementations are not able to recognize any distributional changes (e.g., change of variance), even if these can be related to covariates \citep{Hothorn.2021}. As such, basic versions of XGBoost and LightGBM don't provide a way to model the full predictive distribution $F_{Y}(y|\mathbf{x})$, as they focus on predicting the conditional mean $\mathbb{E}(Y|\mathbf{X} = \mathbf{x})$ only.

\begin{figure}[h!]
	\centering
	\begin{subfigure}{0.6\textwidth}
		\centering
		\caption{Unconditional Density Plot}
		\includegraphics[width=1.0\linewidth]{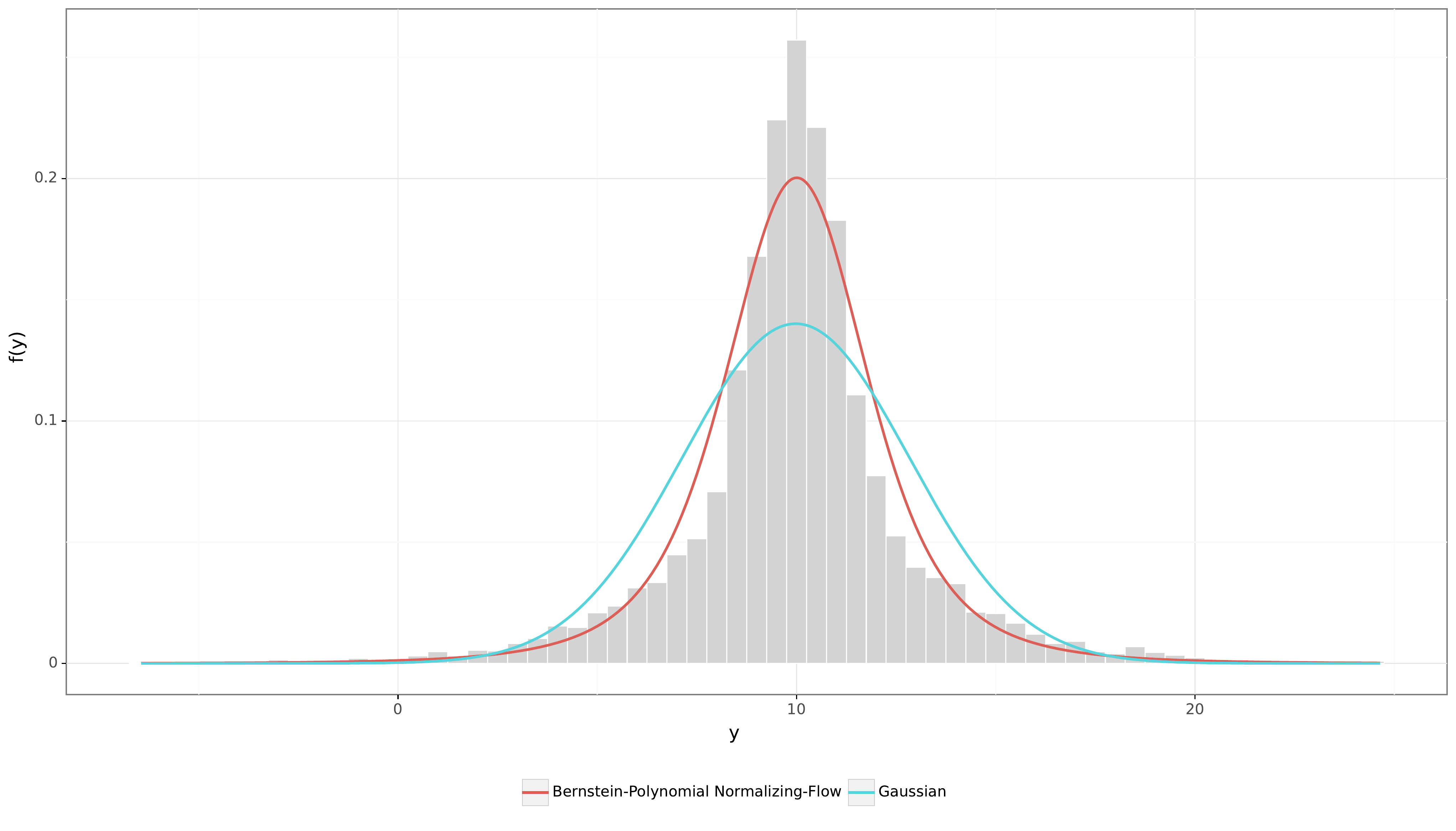}
		\label{fig:gbmlss_xgb_sim}
	\end{subfigure}

	\begin{subfigure}{0.6\textwidth}
		\centering
		\caption{Unconditional CDF Plot}
		\includegraphics[width=1.0\linewidth]{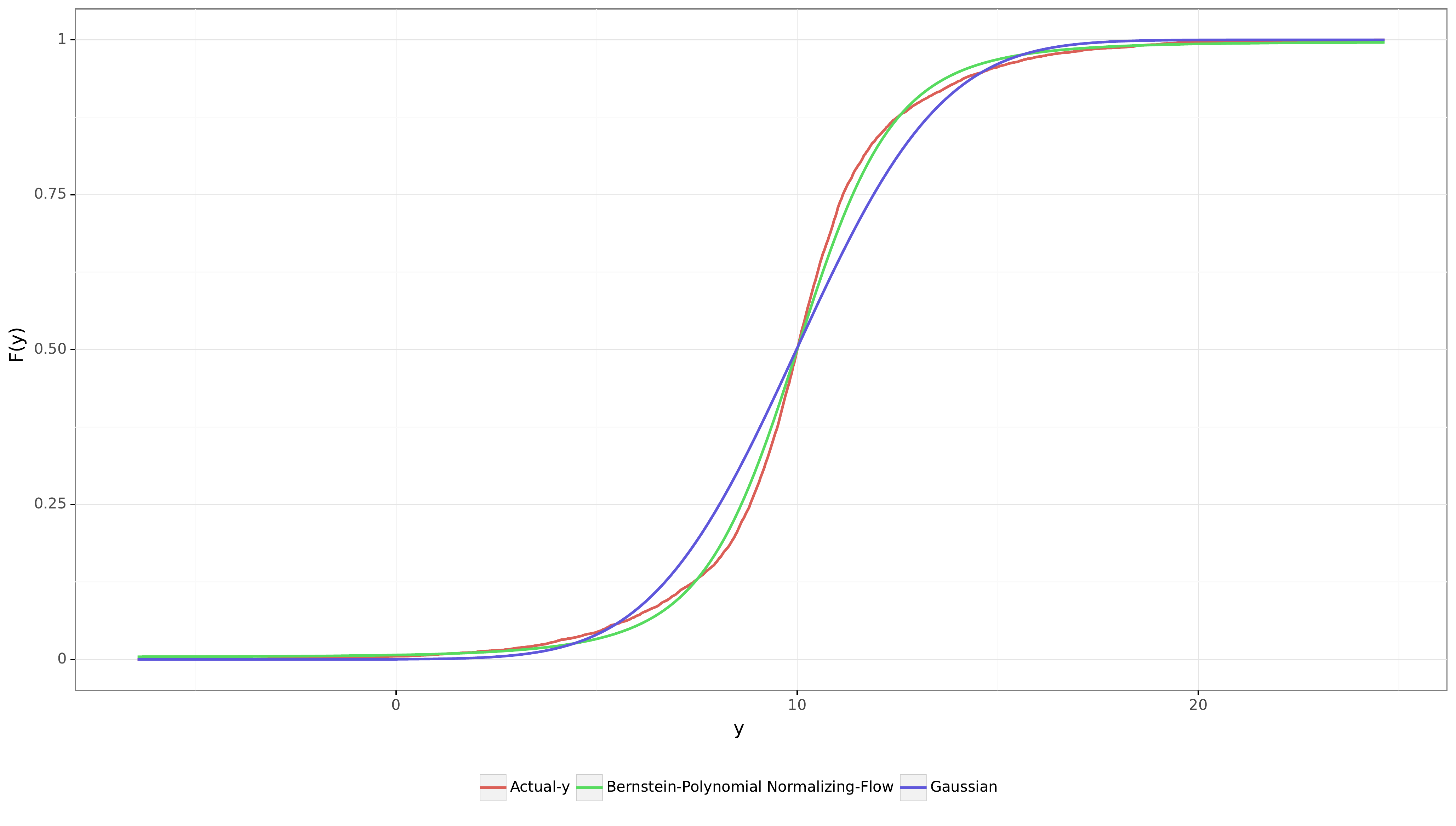}
		\label{fig:gbmlss_lgb_sim}
	\end{subfigure}
	\caption{Unconditional density and CDF plots of the Gaussian and Bernstein-Polynomial Normalizing-Flow.}
	\label{fig:sim_data_hist}
\end{figure}

In general, the syntax of our DGBM models is similar to the original XGBoost and LightGBM implementations. However, the user has to make a distributional assumption for GBMLSS by specifying a family in the function call, as well as to specify the order of the Bernstein-Polynomial $M$ for NFBoost. Since the data have been generated by a Normal distribution, we use the Normal as an input for GBMLSS. For NFBoost, we set $M=6$.\footnote{We ave also varied the polynomial order of NFBoost and found $M=6$ to provide a good approximation to the simulated data.} Since our framework is likelihood-based, we can use unconditional density and CDF estimates to compare and evaluate the distributional fit prior to estimating the models.\footnote{NLL-scores can be used to select an appropriate distribution from a variety of continuous, discrete or mixed discrete-continuous response for GBMLSS and to specify the order of the Bernstein-Polynomial $M$ for NFBoost.} Figure \ref{fig:sim_data_hist} shows that the Bernstein-Polynomials slightly better capture the kurtotic shape of the data compared to the Normal assumption. 

\noindent The ability of the Bernstein-Polynomial Flow to better approximate the data is also confirmed by the likelihood comparison shown in Table \ref{tab:sim_data_nll}.

\begin{table}[h!]
	\begin{center}
		\scalebox{0.9}{
			\begin{threeparttable}
				\caption{NegLogLikelihood Comparison of Gaussian and Bernstein-Polynomial Normalizing-Flow}
				\begin{tabular}{rrrrrrrr}
					\toprule
					Distribution     & NLL \\   
					\midrule
					Gaussian  & 6.9935 \\  
					Bernstein-Polynomial Normalizing-Flow ($M$=6)       & 2.3688\\ 
					\bottomrule
				\end{tabular}
%				\begin{tablenotes}
%					\scriptsize
%					\item \hspace{-0.7em} 
%				\end{tablenotes}
				\label{tab:sim_data_nll}
		\end{threeparttable}
	}
	\end{center}
\end{table}

\begin{figure}[h!]
	\centering
	\begin{subfigure}{.4\textwidth}
		\centering
		\caption{GBMLSS-XGB}
		\includegraphics[width=1.0\linewidth]{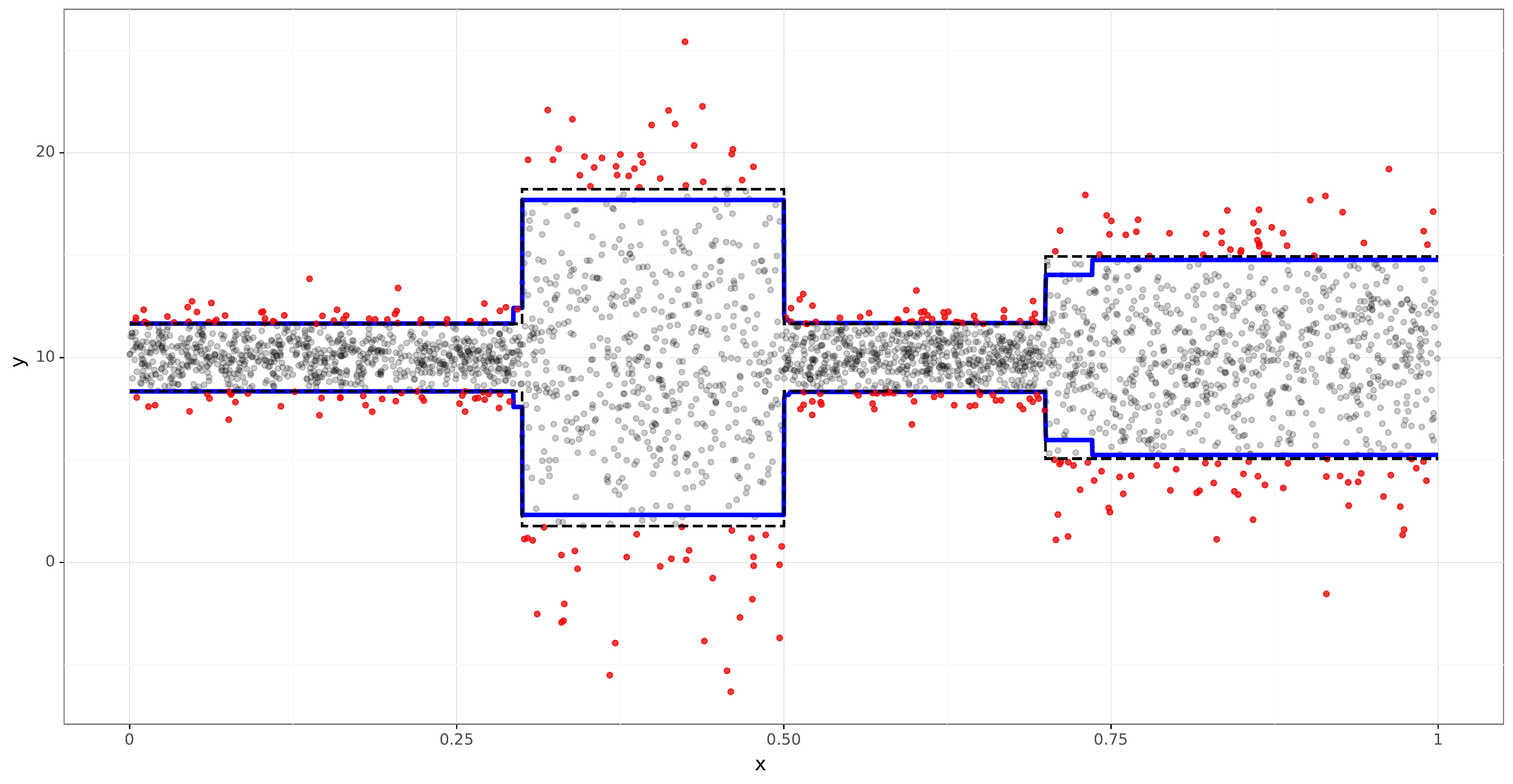}
		\label{fig:gbmlss_xgb_sim}
	\end{subfigure}%
	\begin{subfigure}{0.4\textwidth}
		\centering
		\caption{GBMLSS-LGB}
		\includegraphics[width=1.0\linewidth]{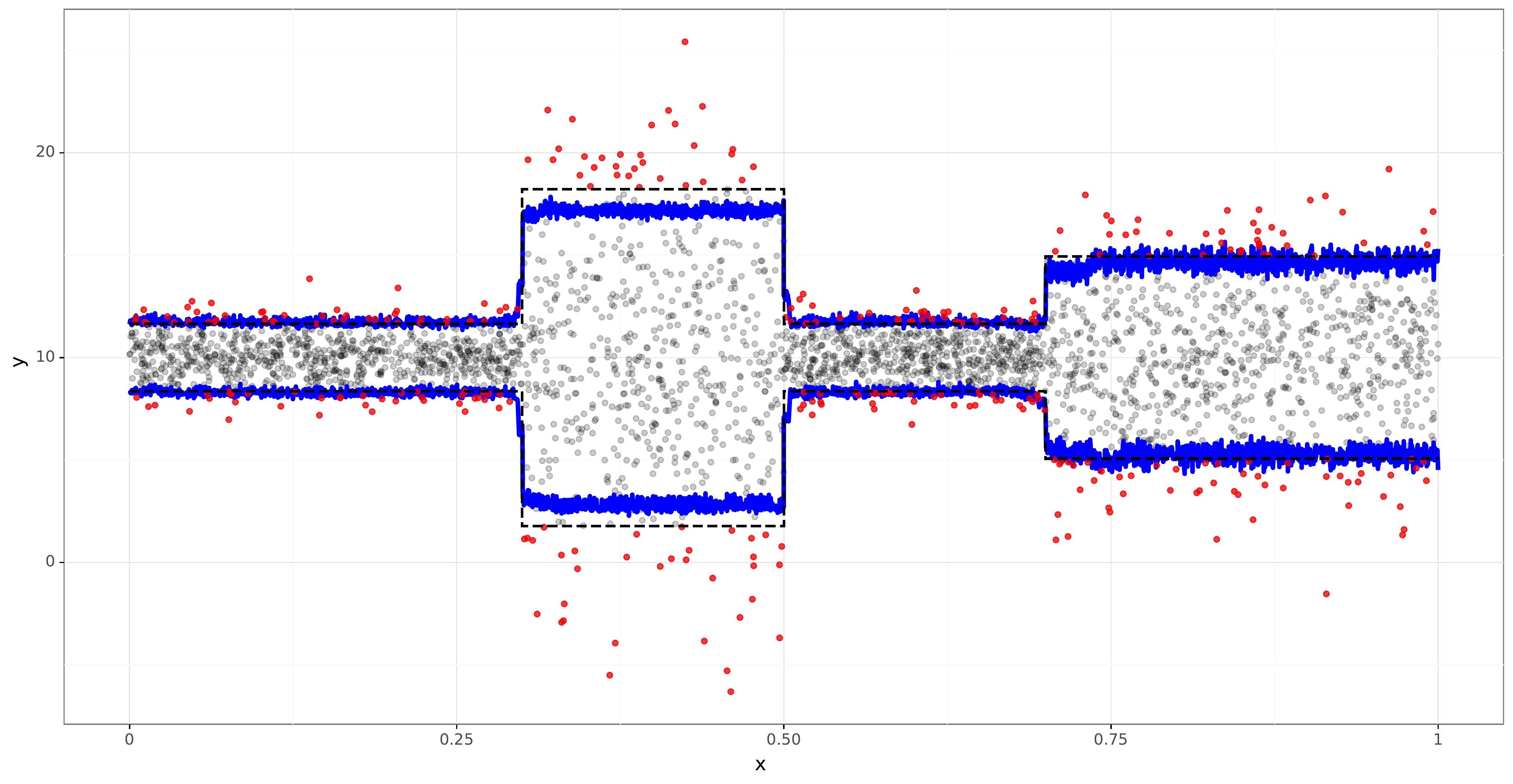}
		\label{fig:gbmlss_lgb_sim}
	\end{subfigure}
	\begin{subfigure}{.4\textwidth}
		\centering
		\caption{NFBoost-XGB}
		\includegraphics[width=1.0\linewidth]{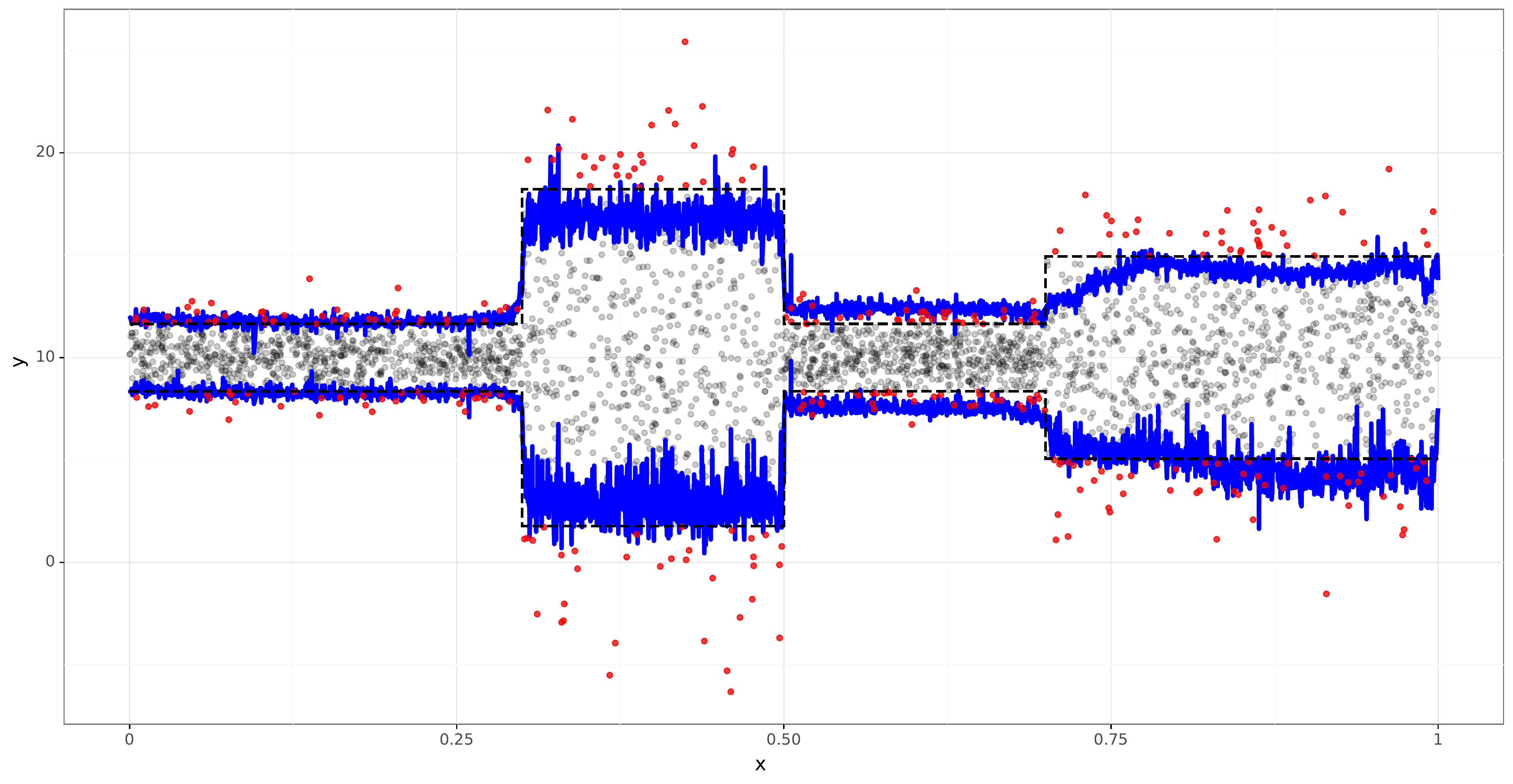}
		\label{fig:nfboost_xgb_sim}
	\end{subfigure}%
	\begin{subfigure}{0.4\textwidth}
		\centering
		\caption{NFBoost-LGB}
		\includegraphics[width=1.0\linewidth]{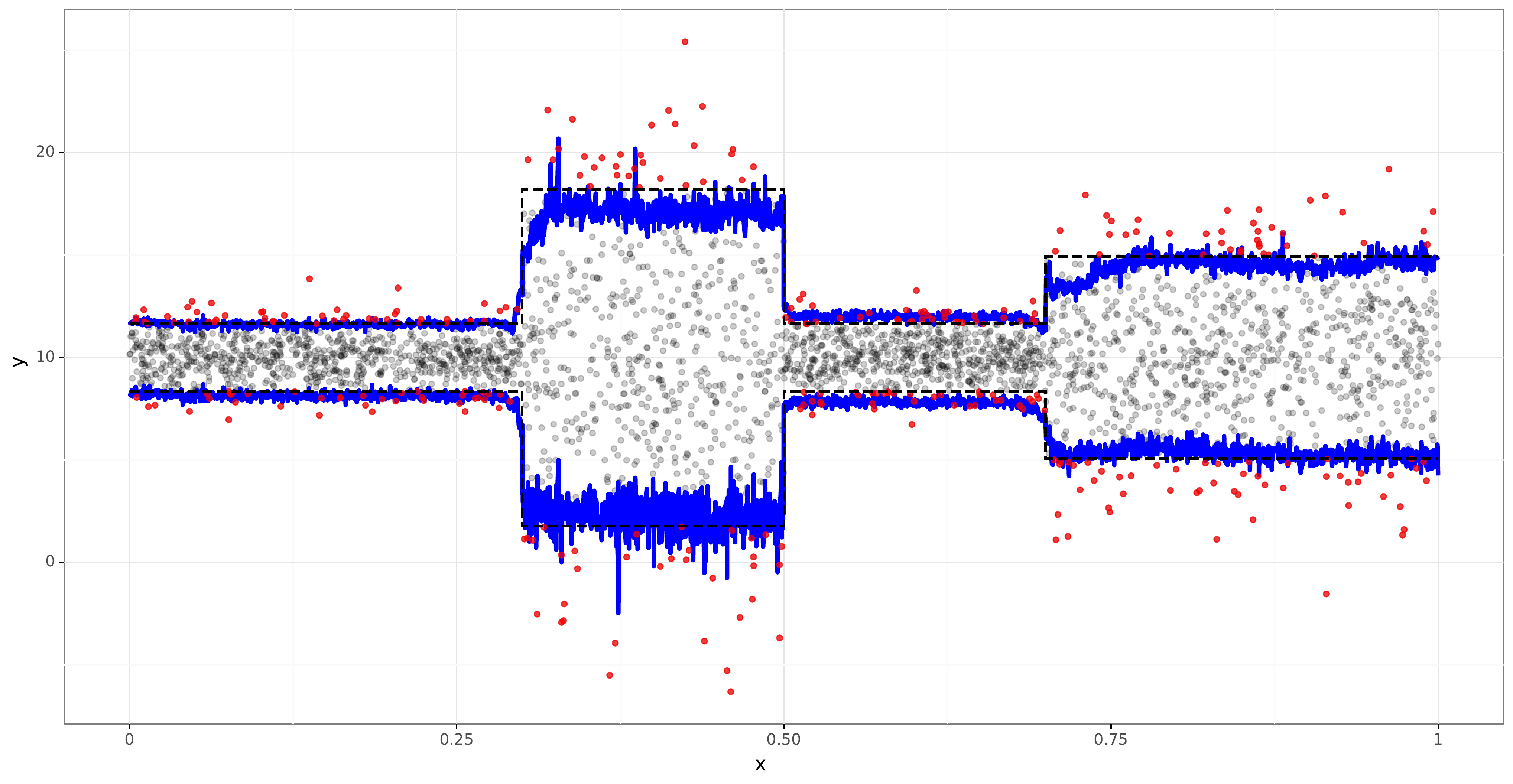}
		\label{fig:nfboost_lgb_sim}
	\end{subfigure}
	\caption{Simulated Test Dataset with 3,000 observations $y \sim \mathcal{N}(10,(1 + 4(0.3 < x < 0.5) + 2(x > 0.7))$. Points outside the conditional 5\% and 95\% quantile are in red. The black dashed lines depict the actual 5\% and 95\% quantiles. Conditional 5\% and 95\% quantile predictions obtained from the DGBM models are depicted by the blue lines. Besides the only informative predictor $x$, we have added $X_{1}, \ldots, X_{10}$ as noise variables.}
	\label{fig:sim_data_fcst}
\end{figure}

\noindent When fitting both models, the user has the option of providing a list of hyper-parameters to find an optimized set of parameters using Optuna of \cite{Akiba.2019}. Once the model is trained, we obtain prediction intervals and quantiles of interest directly from the predicted distribution. Figure \ref{fig:sim_data_fcst} shows the predictions of the DGBM models for the 5\% and 95\% quantile in blue.

\noindent Investigation of Figure \ref{fig:sim_data_fcst} shows that all models in our DGBM framework correctly estimate the heteroskedasticity in the data. The top panels of Figure \ref{fig:sim_data_fcst} further show that GBMLSS predictions of the 5\% and 95\% quantiles are less variable and show a higher alignment with the theoretical quantiles. The difference in the level of variability between NFBoost and GBMLSS predictions becomes  most apparent when comparing Panel \ref{fig:gbmlss_xgb_sim} with Panel \ref{fig:nfboost_xgb_sim}: while both models use XGBoost as a computational engine, Panel \ref{fig:nfboost_xgb_sim} shows a much higher variability of the predicted quantiles. However, comparing Panel \ref{fig:nfboost_xgb_sim} with Panel \ref{fig:nfboost_lgb_sim} also shows that while using the same polynomial order $M$, LightGBM predictions exhibit less variability. One reason might be that a parametric distributional assumption with less parameters to estimate leads to a more stable prediction: for the Gaussian, the GBMLSS models need to estimate 2 parameters only, whereas for the Normalizing-Flow based models, $M+$4 parameters need to be estimated.\footnote{Increasing the data size might be one way to improve the informational content needed to estimate a larger number of parameters. Another reason for the lower variability might be that for the simulation data example, all GBMLSS models are trained using exact gradients and hessians, whereas gradients and hessians for the NFBoost models are approximated using automatic differentiation. To further stabilize the estimation, one can also run a more exhaustive hyper-parameter search.} Its smooth approximation of the theoretical quantiles shown in Panel \ref{fig:nfboost_xgb_sim} also indicates that GBMLSS-XGB profits the most from the unconditional parameter initialization: while all other models iterate to, or close to the maximum number of 500 boosting rounds for all hyper-parameter combinations, GBMLSS-XGB stops already early after 25 iterations.    

\begin{figure}[h!]
	\centering
	\begin{subfigure}{.4\textwidth}
		\centering
		\caption{GBMLSS-XGB Feature Importance}
		\includegraphics[width=1.0\linewidth]{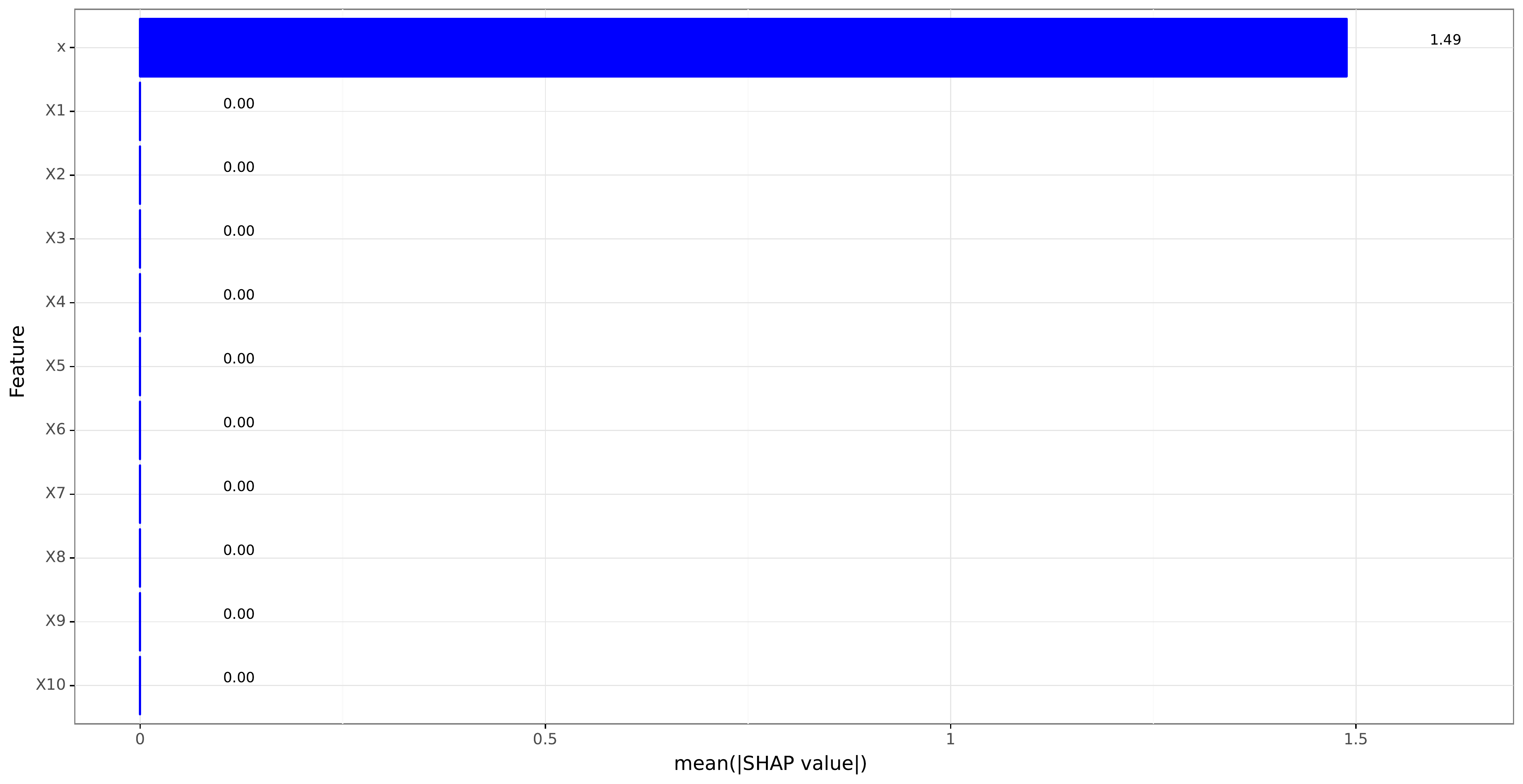}
		\label{fig:gbmlss_xgb_imp}
	\end{subfigure}%
	\begin{subfigure}{0.4\textwidth}
		\centering
		\caption{GBMLSS-XGB Partial Dependence Plot}
		\includegraphics[width=1.0\linewidth]{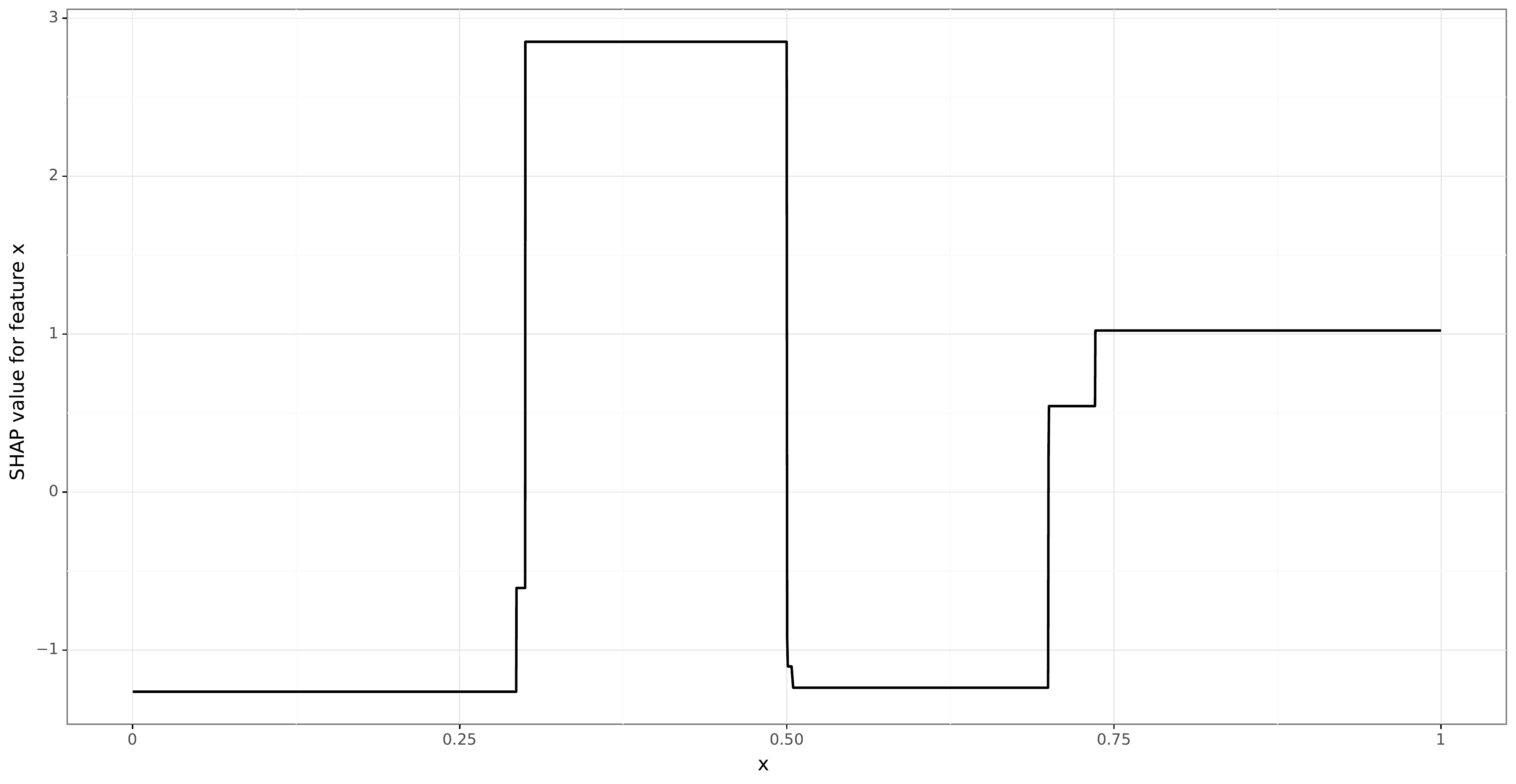}
		\label{fig:gbmlss_xgb_pdp}
	\end{subfigure}
	\begin{subfigure}{.4\textwidth}
		\centering
		\caption{GBMLSS-LGB Feature Importance}
		\includegraphics[width=1.0\linewidth]{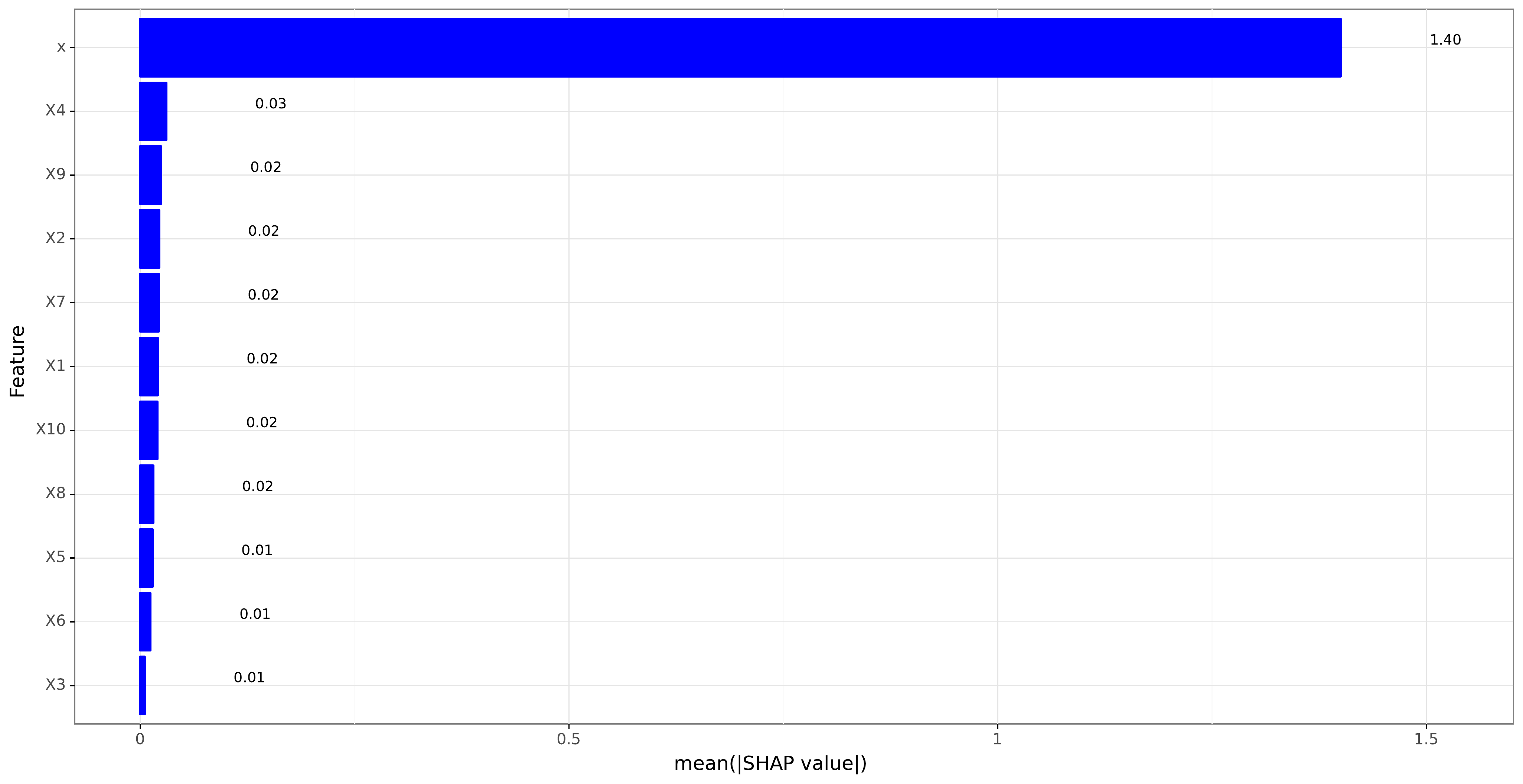}
		\label{fig:gbmlss_lgb_imp}
	\end{subfigure}%
	\begin{subfigure}{0.4\textwidth}
		\centering
		\caption{GBMLSS-LGB Partial Dependence Plot}
		\includegraphics[width=1.0\linewidth]{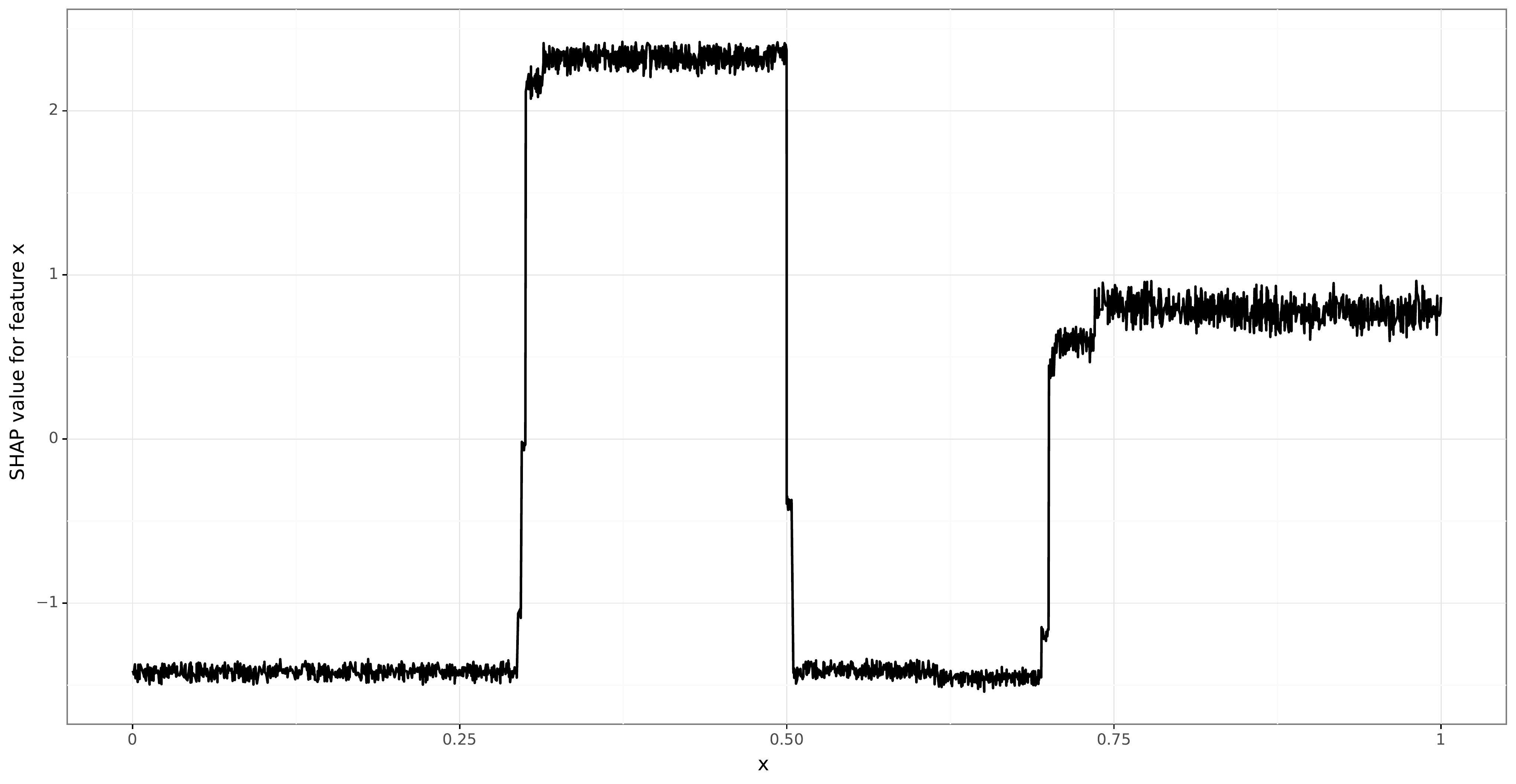}
		\label{fig:gbmlss_lgb_pdp}
	\end{subfigure}
	\caption{Feature importance and partial dependence plots using Shapley Values.}
	\label{fig:sim_data_imp}
\end{figure}

Our GBMLSS models also provide insights into the data generating process via feature importances and partial dependence plots for all distributional parameters.\footnote{While partial dependence plots and attribute importances are also available for NFBoost models, Bernstein-Polynomial coefficients don't have a direct interpretation, which renders any analysis of them difficult.} Since the conditional mean is simulated as being constant, Figure \ref{fig:sim_data_imp} displays the effect on the conditional variance $\mathbb{V}(Y|\mathbf{X} = \mathbf{x})$ only. 

\noindent The top panels of Figure \ref{fig:sim_data_imp} show that all GBMLSS models correctly identify the only informative predictor $x$ and do not consider any of the noise variables $X_{1}, \ldots, X_{10}$ as important features. The partial dependence plots in the right panels of Figure \ref{fig:sim_data_imp} also show that both models correctly identify the shape and magnitude of the heteroskedasticity in the data. As already suggested by the predicted quantiles shown in Figure \ref{fig:sim_data_fcst}, Panels \ref{fig:gbmlss_xgb_pdp} and \ref{fig:gbmlss_lgb_pdp} confirm that the conditional variance estimate of GBMLSS-XGB appears to be less variable compared to the GBMLSS-LGB estimate. 

\subsection{UCI Regression Datasets} \label{sec:uci} 

In this section, we benchmark our DGBM framework against NGBoost \citep{Duan.2020} and PGBM \citep{Sprangers.2021} using a subset of the UCI-datasets of \cite{Dua.2017}. For conducting the experiments, we proceed as follows: for each dataset, we create 5 randomly shuffled folds and split each fold into train and test, where we keep 10\% of the data for evaluating the models and use the remaining 90\% for training. To make the results comparable to \citet{Sprangers.2021, Duan.2020}, we assume a Gaussian distribution for all parametric distributional models and keep the set of hyper-parameters constant across all datasets.\footnote{One exception is the yacht dataset, where we increase the learning rate of all NFBoost models to 0.03. For NGBoost and PGBM, we use the same set of hyper-parameters as reported in \citet{Sprangers.2021, Duan.2020} and configure all GBMLSS and NFBoost models to have similar settings as PGBM. Due to its highly skewed nature, we log-transform the yacht response. To stabilize estimation of gradients and hessians, we scale all responses $y/100$.} As such, neither early stopping nor hyper-parameter tuning is used during the experiments. All models are trained based on 1,000 boosting rounds. Table \ref{tab:uci_params} summarises the hyper-parameters used for conducting the experiments.

\begin{table}[h!]
	\begin{center}
		\scalebox{0.8}{
			\begin{threeparttable}
				\caption{Hyper-Parameters for UCI-dataset experiments}
				\begin{tabular}{lcccccc}
					\toprule
					& GBMLSS-LGB &               GBMLSS-XGB &                NFBoost-LGB &                 NFBoost-XGB &                  NGBoost &                       PGBM \\ 
					\midrule
					$\mbox{min\_split\_gain (gamma)}$   &  0  &  0  &  0  &  0  &  0  &  0 \\ 
					$\mbox{min\_data\_in\_leaf}$   &  1  &  1  &  1  &  1  &  1  &  1 \\ 
					$\mbox{min\_data\_in\_bin}$   &  1  &  na  &  1  &  na  &  na  &  na \\ 
					$\mbox{max\_bin}$   &  64  &  64  &  64  &  64  &  na  &  64 \\ 
					$\mbox{max\_leaves}$   &  16  &  16  &  16  &  16  &  na  &  16 \\ 
					$\mbox{max\_depth}$   &  -1  &  0  &  -1  &  0  &  3  &  -1 \\ 
					$\mbox{learning\_rate}$   &  0.1  &  0.1  &  0.01/0.03  &  0.01/0.03  &  0.01  &  0.1 \\ 
					$\mbox{boosting\_rounds}$   & 1,000 &  1,000  &  1,000  &  1,000  &  1,000  &  1,000 \\ 
					$\mbox{n\_data\_folds}$   & 5 &  5  &  5  &  5  &  5  &  5 \\ 
					$\mbox{lambda (alpha)}$   & 1.0 &  1.0  &  1.0  &  1.0  &  na  &  1.0 \\ 
					Bernstein-Order $M$ / Distribution   & Gaussian &  Gaussian  &  \{10, 4, 8, 4, 10, 3\}  &  \{10, 4, 8, 4, 10, 3\}  &  Gaussian  &  Gaussian \\ 
					\bottomrule
				\end{tabular}
				\begin{tablenotes}
					\scriptsize
					\item \hspace{-0.7em} For the yacht dataset, we log-transform the response and increase the learning rate of all NFBoost models from 0.01 to 0.03. To stabilize estimation of gradients and hessians, we scale all responses $y/100$. The table only shows hyper-parameter settings that deviate from default values. To use similar settings for both XGB and LGB models, we set $tree\_method = hist$ and $grow\_policy = lossguide$ for XGB based models. The dataset order of Bernstein-Polynomials $M$ is \{boston, concrete, kin8nm, naval, protein, yacht\}.
				\end{tablenotes}
				\label{tab:uci_params}
		\end{threeparttable}
	}
	\end{center}
\end{table}

\noindent Probabilistic forecasts of all models are evaluated using the Continuous Ranked Probability Score (CRPS) of \cite{Gneiting.2007}, which is a measure of the difference between the predicted and the empirical cumulative distribution function of the ground-truth observation $y$

\begin{equation}
CRPS(F,y)=\int_{\mathbb{R}}\big(F(z)-\mathds{I}\{y \leq z\}\big)^{2}dz
\end{equation}

\noindent where $\mathds{I}\{y \leq z\}$ denotes the indicator function. The CRPS is a proper scoring function that attains a minimum if the predicted distribution $F$ and the data distribution are identical.\footnote{Following \citet{Rangapuram.2021, Kurle.2020}, the CRPS can be re-formulated as follows
	
\begin{equation}
	CRPS(\hat{F},y)=\int^{1}_{0} QL_{\alpha}\Big(\hat{F}^{-1}, y\Big)d\alpha \nonumber 
\end{equation}

\noindent where $QL_{\alpha} = 2\big(\mathds{I}\{y \leq \hat{F}^{-1}(\alpha)\}-\alpha\big)\big(\hat{F}^{-1}(\alpha)-y\big)$ denotes the $\alpha$-quantile loss evaluated at $\alpha \in [0,1]$. From the above equation it follows that the CRPS can be interpreted as the quantile loss integrated over all quantile levels $\alpha \in [0,1]$ \citep{Gasthaus.2019}.} To evaluate all models, we use 1,000 samples drawn from the predicted distributions.

To determine the order of the Bernstein-Polynomial, we search over the grid $M \in [3,10]$ and select the order with the lowest negative log-likelihood. We want to stress that the selection of $M$ is crucial since it determines the flexibility of the estimated cumulative distribution function. In fact, very low and very high orders of $M$ represent two different extremes: for $M$=1 and $F_{Z} = N(0,1)$, $\hat{F}_{Y}$ belongs to the family of Gaussian distribution functions, whereas for $M=n-1$, $\hat{F}_{Y}$ constitutes an interpolation of the data. Therefore, since Bernstein-Polynomial Flows can approximate distribution functions of exquisite complexity for high $M$, an appropriate choice of $M$ is crucial to maintain a balance between a good approximation and the ability of the model to extrapolate beyond unseen data. 

\begin{figure}[h!]
	\begin{subfigure}{.5\textwidth}
		\centering
		\caption{}
		\includegraphics[width=1.0\linewidth]{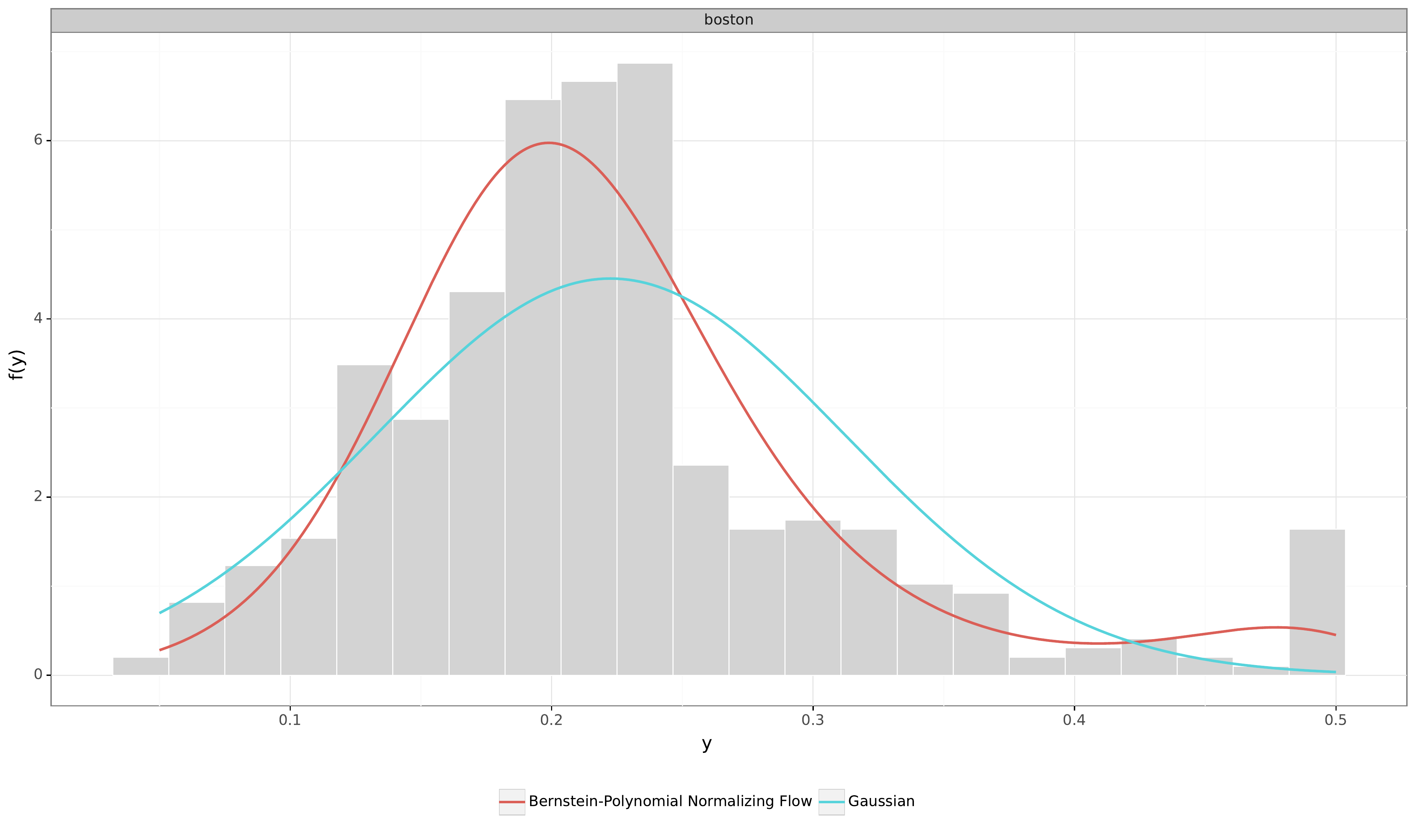}
		\label{fig:uci_boston}
	\end{subfigure}
	\begin{subfigure}{.5\textwidth}
		\centering
		\caption{}
		\includegraphics[width=1.0\linewidth]{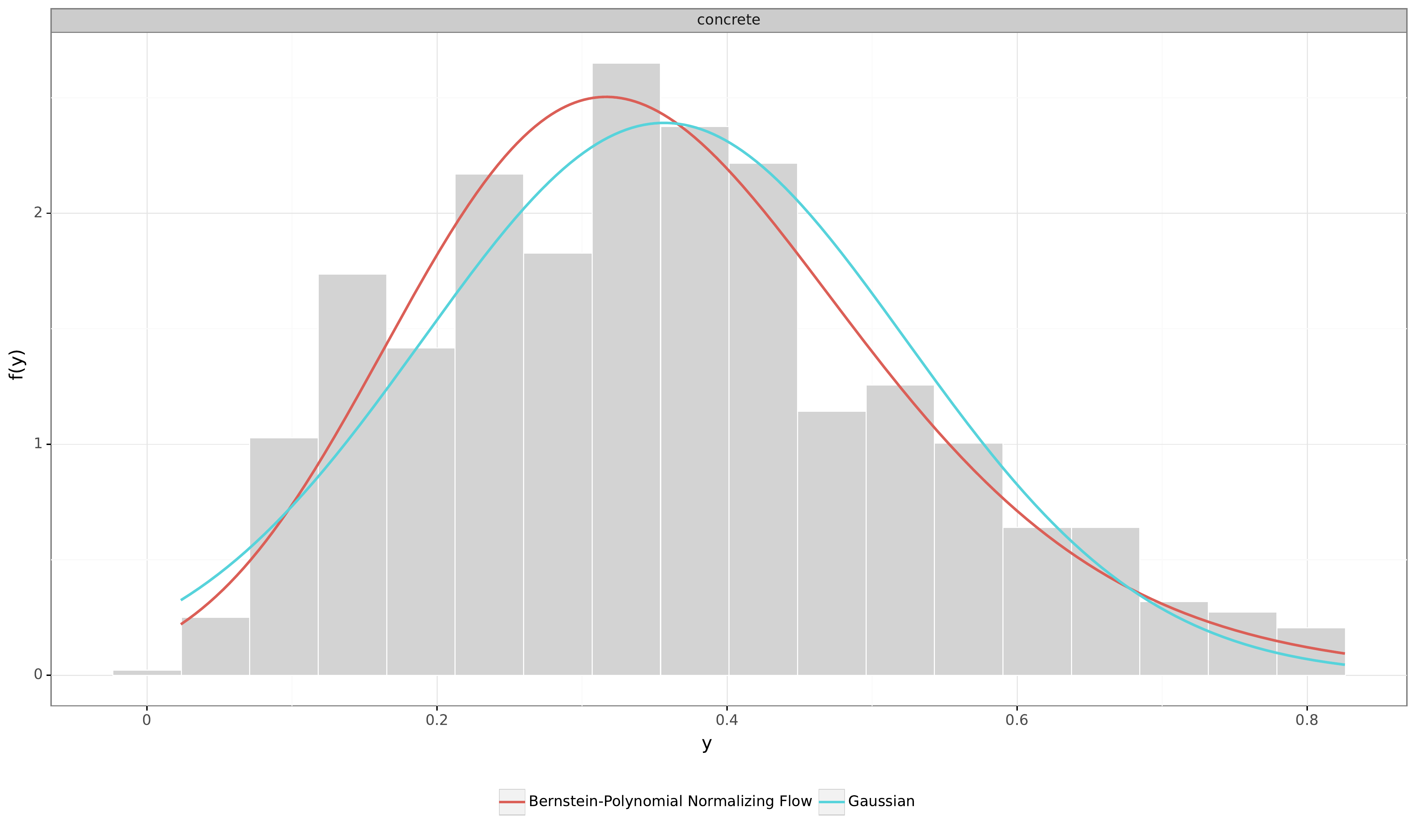}
		\label{fig:uci_concrete}
	\end{subfigure}
	\begin{subfigure}{.5\textwidth}
		\centering
		\caption{}
		\includegraphics[width=1.0\linewidth]{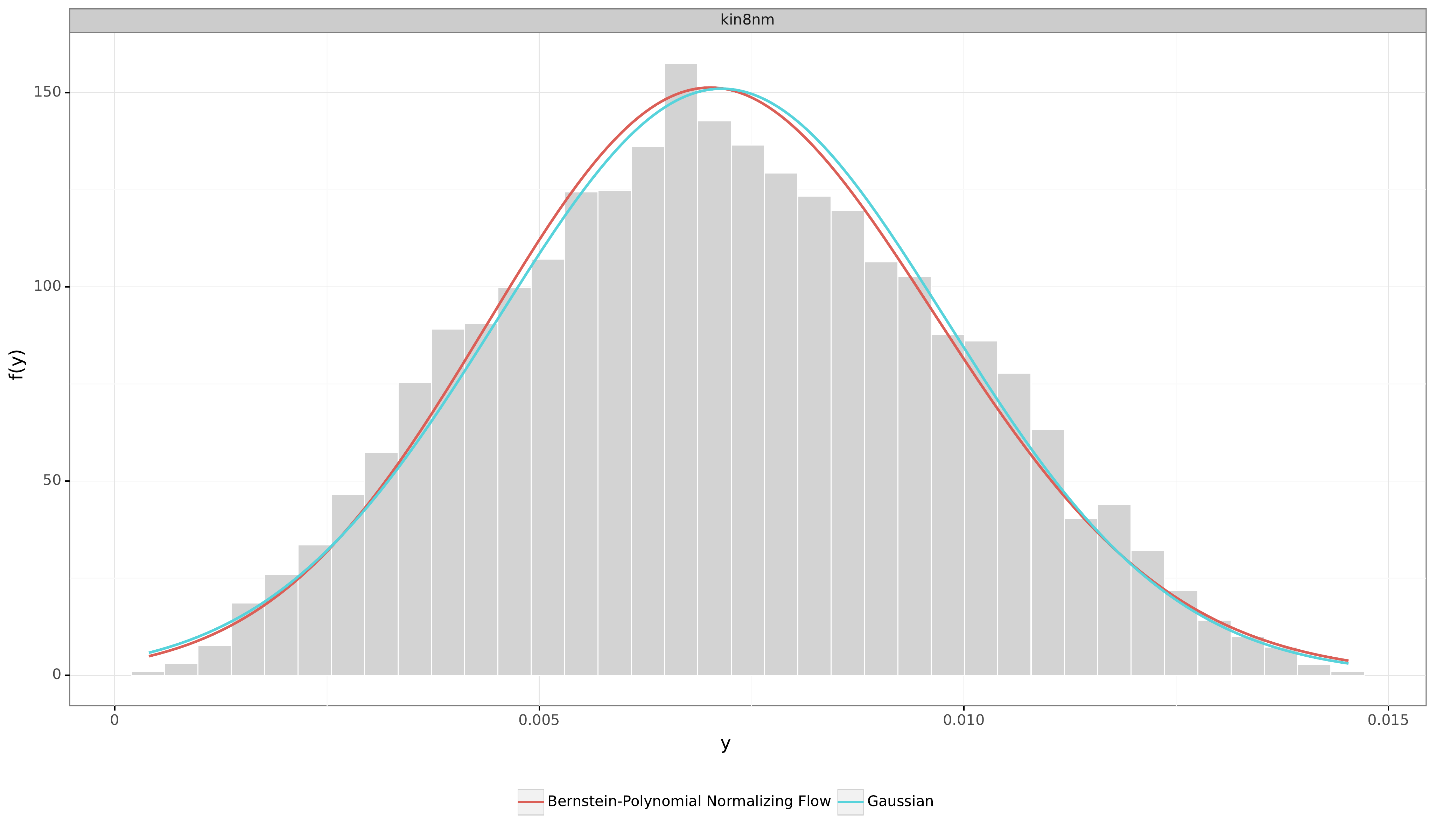}
		\label{fig:uci_kin8nm}
	\end{subfigure}
	\begin{subfigure}{.5\textwidth}
		\centering
		\caption{}
		\includegraphics[width=1.0\linewidth]{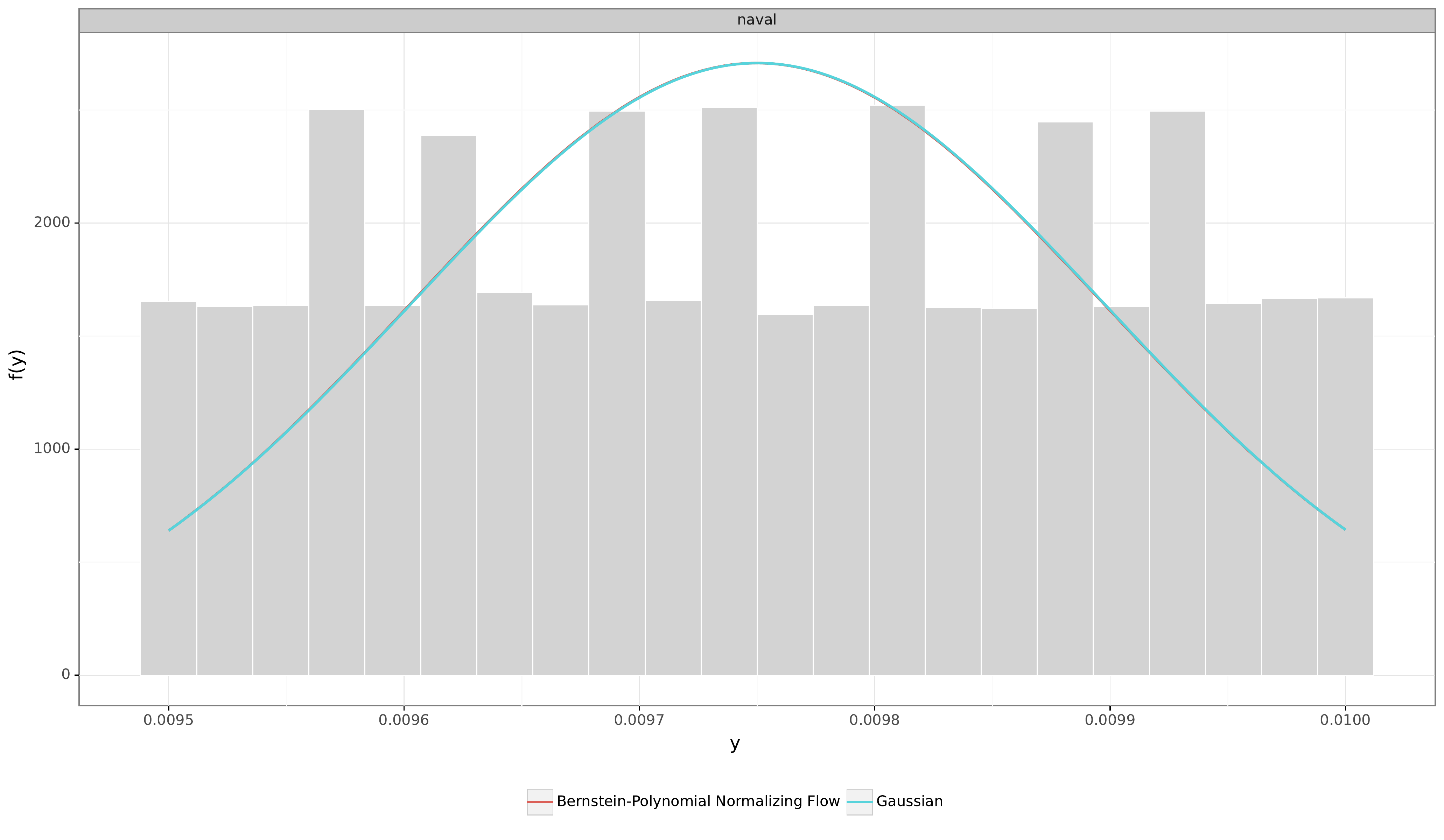}
		\label{fig:uci_naval}
	\end{subfigure}
	\begin{subfigure}{.5\textwidth}
		\centering
		\caption{}
		\includegraphics[width=1.0\linewidth]{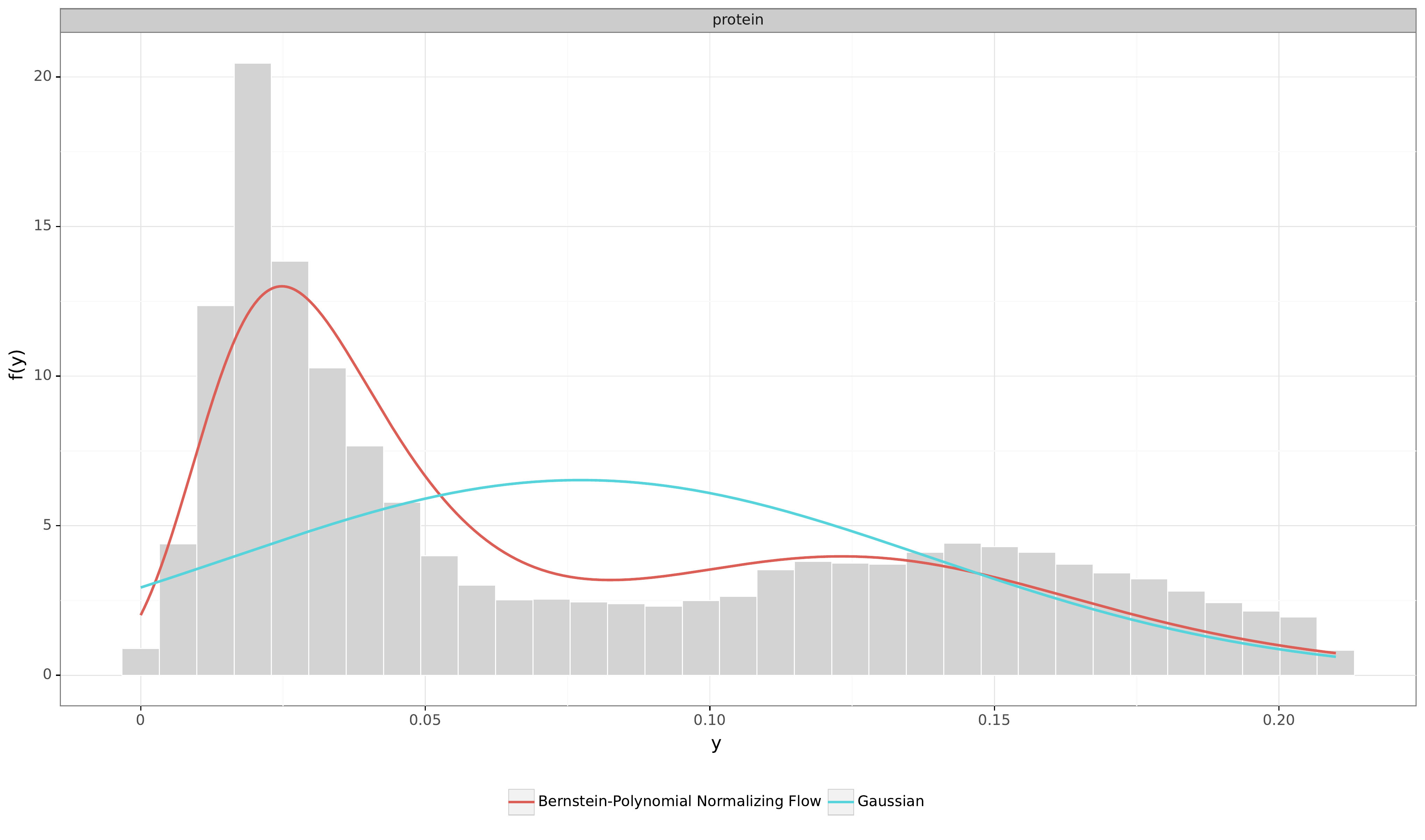}
		\label{fig:uci_protein}
	\end{subfigure}
	\begin{subfigure}{.5\textwidth}
		\centering
		\caption{}
		\includegraphics[width=1.0\linewidth]{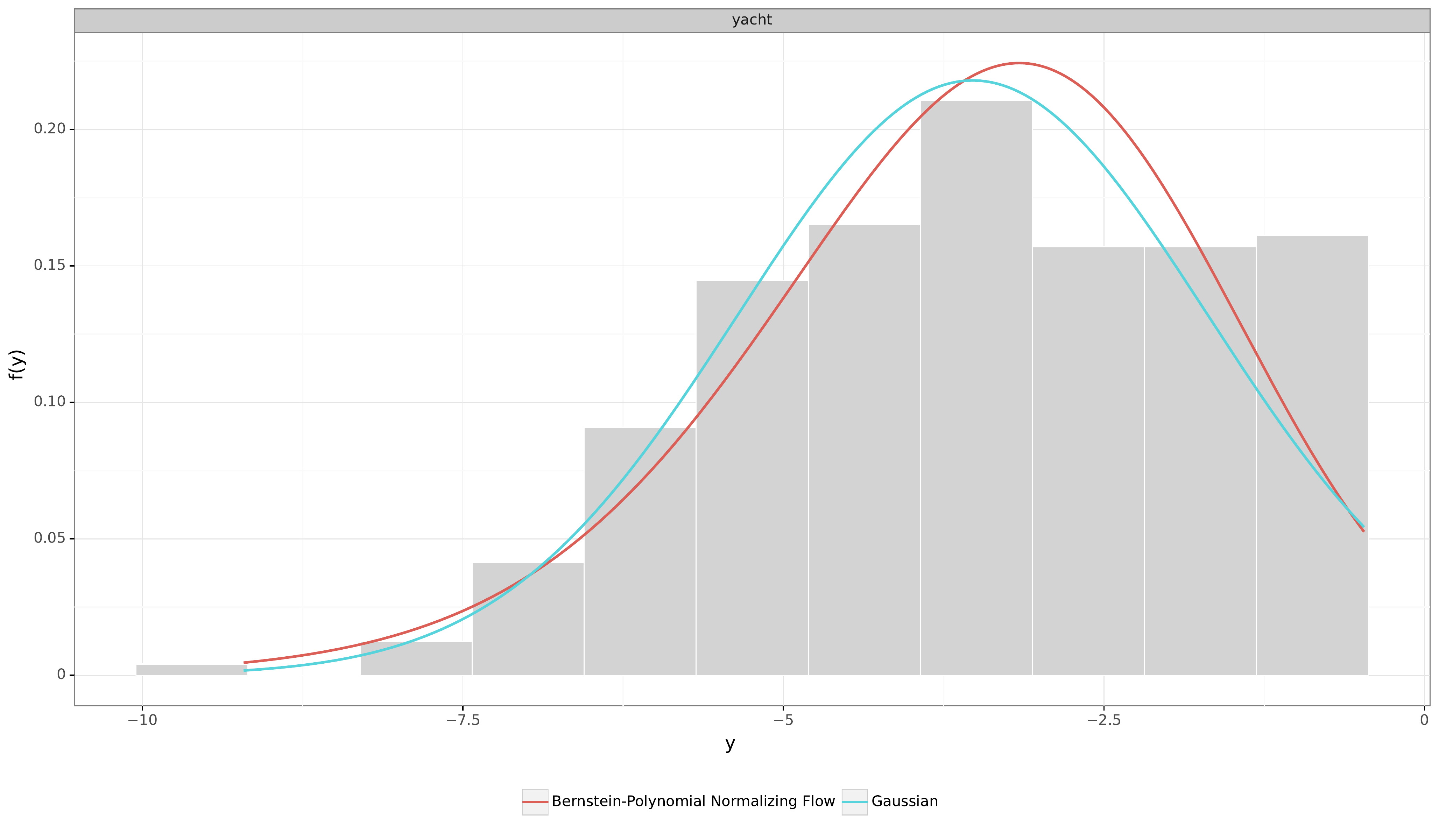}
		\label{fig:uci_yacht}
	\end{subfigure}
	\caption{Unconditional density plots of the Gaussian and Bernstein-Polynomial Normalizing-Flow for the UCI-Datasets.}
	\label{fig:uci_bp_order}
\end{figure}

Before we present the results of the experiments, we investigate Figure \ref{fig:uci_bp_order} that shows unconditional density plots of the Gaussian and Bernstein-Polynomial Normalizing-Flow for the UCI-Datasets. 

\noindent The ability of the Bernstein-Polynomial Flow to reconstruct any data-generating process is most evident from Panel \ref{fig:uci_protein}, where we expect the NFBoost models to have a higher accuracy than parametric distribution models, since the protein dataset exhibits a bimodal behaviour that is not well captured by the Gaussian. We now turn to the discussion of the CRPS scores presented in Table \ref{tab:uci_crps_scores}.

%\noindent For some datasets, e.g., kin8nm and concrete, the Gaussian provides a reasonable approximation, even though the Bernstein-Polynomial Flow shows that the concrete data are slightly skewed. The ability of the Bernstein-Polynomial Flow to reconstruct any data-generating process is most evident from Panel \ref{fig:uci_protein}, where we expect the NFBoost models to have a higher accuracy than parametric distribution models, since the protein dataset exhibits a bimodal behaviour that is not well captured by the Gaussian. Even though both the Bernstein-Polynomial Flow and the Gaussian fit show close agreement for the naval dataset of Panel \ref{fig:uci_naval}, the data seem to be rather uniformly distributed. After the log-transformation, the yacht data of Panel \ref{fig:uci_yacht} still exhibit some skewness that deviates from the Gaussian, as indicated by the Flow fit. We now turn to the discussion of the CRPS scores presented in Table \ref{tab:uci_crps_scores}.

\begin{table}[h!]
	\begin{center}
		\scalebox{0.7}{
			\begin{threeparttable}
				\caption{CRPS scores across models and UCI datasets}
				\begin{tabular}{lcccccc}
					\toprule
					& GBMLSS-LGB &               GBMLSS-XGB &                NFBoost-LGB &                 NFBoost-XGB &                  NGBoost &                       PGBM \\ 
					\midrule
					boston (N=506)   &    1.6935 [1.5855, 1.73] &   1.6153 [1.5198, 1.629] &    1.9368 [1.6219, 2.0662] &     \textbf{1.5549 [1.5537, 1.5746]} &  1.7007 [1.6143, 1.8061] &    1.5715 [1.5664, 1.5991] \vspace{1em} \\ 
					concrete (N=1,030) &  \textbf{1.6172 [1.5444, 1.7071]} &   1.642 [1.5611, 1.8968] &    2.1467 [1.9287, 2.2436] &     2.3532 [2.0562, 2.3712] &   2.5161 [2.3637, 2.682] &    1.8121 [1.6745, 2.2021] \vspace{1em} \\ 
					kin8nm (N=8,192)   &   0.0729 [0.0728, 0.073] &   0.073 [0.0724, 0.0731] &    \textbf{0.0629 [0.0625, 0.0632]} &     0.0639 [0.0625, 0.0641] &  0.0965 [0.0941, 0.0968] &     0.1401 [0.1396, 0.195] \vspace{1em} \\ 
					naval (N=11,934)    &  0.0034 [0.0034, 0.0034] &  0.0034 [0.0034, 0.0034] &    0.0035 [0.0035, 0.0035] &     0.0035 [0.0035, 0.0035] &  \textbf{0.0017 [0.0017, 0.0018]} &    0.0061 [0.0057, 0.0069] \vspace{1em} \\ 
					protein (N=45,730)  &  1.9898 [1.9839, 2.0063] &  1.9974 [1.9869, 2.0034] &    \textbf{1.9031 [1.8861, 1.9203]} &      1.935 [1.9245, 1.9378] &  2.5259 [2.5131, 2.5262] &    2.1103 [2.1085, 2.1734] \vspace{1em} \\ 
					yacht (N=308)    &  \textbf{5.7119 [5.3285, 7.1834]} &  7.3127 [3.6087, 7.9982] &  10.655 [10.4204, 11.3639] &  25.0991 [22.4859, 25.8206] &  6.1083 [5.6999, 7.4556] &  14.877 [14.1427, 19.3816] \\ 
					\bottomrule
				\end{tabular}
				\begin{tablenotes}
					\scriptsize
					\item \hspace{-0.7em} The table shows median CRPS scores across all folds, with interquartile-range values in parentheses, i.e., $q_{0.5}(q_{0.25}, q_{0.75})$. Lower is better, with best results in bold.
				\end{tablenotes}
				\label{tab:uci_crps_scores}
		\end{threeparttable}}
	\end{center}
\end{table}

\noindent As hypothesized earlier, Normalizing-Flow based models outperform parametric distributional models for the bimodal protein dataset due to their higher flexibility. A similar picture emerges for the slightly skewed and kurtotic boston dataset, where NFBoost-XGB achieves the highest accuracy, closely followed by the PGBM model. The comparatively small number of observations in the boston data set, however,  may account for the unstable behaviour of NFBoost-LGB. From Table \ref{tab:uci_crps_scores} it also appears that for small and medium sized datasets, such as the concrete, where a parametric Gaussian distribution is a reasonable approximation to the data, parametric GBMLSS models are more efficient and provide a higher accuracy than Normalizing-Flow based models. Even though the kin8nm data are also well described by a Gaussian, we attribute the higher accuracy of NFBoost models to the number of observations and hence to the higher informational content in the data, which allows more flexible Flow-based models to be estimated. For the smallest yacht dataset, NFBoost-XGB appears to have the most difficulties, possibly due to the small amount of observations relative to the number of features and parameters being estimated. For the uniformly distributed naval dataset, NGBoost achieves the lowest score, with all models of our DGBM framework scoring equally.\footnote{Despite the favourable results presented in Table \ref{tab:uci_crps_scores}, we want to stress that none of the models used for the experiments are hyper-parameter tuned. Typically, the more flexible a model is, the more sensitive it tends to react to its hyper-parameters. Therefore, a different overall ranking might result with a more exhaustive hyper-parameter search. For example, more parsimonious NFBoost models with shallower trees might be beneficial for smaller datasets, such as the boston and yacht data. We keep the analysis of parameter sensitivity of our framework for future versions of the paper.}  

We also report CRPS ranks in Table \ref{tab:uci_crps_rank}, showing that all models in our DGBM framework achieve state-of-the-art accuracy. Among the DGBM models, GBMLSS achieve the highest accuracy, with GBMLSS-LGB ranking best on average. For the NFBoost models, NFBoost-XGB has a small lead over NFBoost-LGB.

\begin{table}[h!]
	\begin{center}
		\scalebox{0.9}{
			\begin{threeparttable}
				\caption{CRPS score rankings for the UCI datasets}
				\begin{tabular}{lcccccc}
					\toprule
					& GBMLSS-LGB &               GBMLSS-XGB &                NFBoost-LGB &                 NFBoost-XGB &                  NGBoost &                       PGBM \\ 
					\midrule
					boston (N=506)   &           4 &           3 &            6 &            1 &        5 &     2 \\
					concrete (N=1,030) &           1 &           2 &            4 &            5 &        6 &     3 \\
					kin8nm (N=8,192)   &           3 &           4 &            1 &            2 &        5 &     6 \\
					naval (N=11,934)   &           3 &           2 &            5 &            4 &        1 &     6 \\
					protein (N=45,730)  &           3 &           4 &            1 &            2 &        6 &     5 \\
					yacht (N=308)    &           1 &           3 &            4 &            6 &        2 &     5 \\
					\midrule
					\midrule
					Average Rank &         2.5 &         3.0 &          3.5 &          3.3 &      4.2 &   4.5 \\
					\bottomrule
				\end{tabular}
				\begin{tablenotes}
					\scriptsize
					\item \hspace{-0.7em} Results are ranks based on median CRPS scores reported in Table \ref{tab:uci_crps_scores}. Lower is better.
				\end{tablenotes}
				\label{tab:uci_crps_rank}
		\end{threeparttable}
	}
	\end{center}
\end{table}

In addition to CRPS metrics, runtimes for all models are presented in Table \ref{tab:uci_runtime}.\footnote{All models are CPU-trained on a single machine in a non-distributed manner. For PGBM training, we use the numba implementation.} 

\begin{table}[h!]
	\begin{center}
		\scalebox{0.8}{
		\begin{threeparttable}
			\caption{Average runtime in minutes}
			\begin{tabular}{lcccccc}
				\toprule
				& GBMLSS-LGB &               GBMLSS-XGB &                NFBoost-LGB &                 NFBoost-XGB &                  NGBoost &                       PGBM \\ 
				\midrule
				boston (N=506)   &      \textbf{0.0246} &      0.0436 &       4.4566 &       4.7024 &   0.1166 &  0.0450 \\
				concrete (N=1,030) &      \textbf{0.0206} &      0.0656 &       4.8334 &       5.1640 &   0.1208 &  0.0437 \\
				kin8nm (N=8,192   &      \textbf{0.0432} &      0.3541 &       7.2432 &       7.6311 &   0.6378 &  0.1047 \\
				naval (N=11,934)    &      \textbf{0.0535} &      0.5101 &       5.2801 &       5.9496 &   1.0536 &  0.1560 \\
				protein (N=45,730)  &      \textbf{0.1401} &      1.8121 &      25.8645 &      31.7632 &   3.5939 &  0.4536 \\
				yacht (N=308)     &      \textbf{0.0164} &      0.0347 &       3.9868 &       3.9612 &   0.0927 &  0.0391 \\
				\midrule
				\midrule
				Average Rank &         1.0 &         2.7 &          5.2 &          5.8 &      4.0 &   2.3 \\
				\bottomrule
			\end{tabular}
			\begin{tablenotes}
				\scriptsize
				\item \hspace{-0.7em} Results are runtimes in minutes, averaged across all folds. Lower is better, with minimum runtime in bold. The bottom part shows the runtime rank averaged across all datasets.
			\end{tablenotes}
			\label{tab:uci_runtime}
		\end{threeparttable}
	}
	\end{center}
\end{table}

\noindent Combining CRPS scores and runtime ranks of Table \ref{tab:uci_runtime} shows that GBMLSS-LGB is not only the fastest, but also the most accurate model on the benchmark datasets, with PGBM being close second. This corresponds with the findings of the M5 forecasting competition, where LightGBM consistently scores high, both for the point and probabilistic competitions. However, Table \ref{tab:uci_runtime} also shows that runtimes for NFBoost models are not yet competitive, constantly exceeding those of other approaches by several orders of magnitude, especially for larger datasets such as the protein. Since the analysis and efficiency improvements are still ongoing, we can only formulate some mild hypotheses about the reasons for the runtime issue at this stage. One reason might be the known fact that XGBoost and LightGBM scale $\mathcal{O}(P^{2})$ with the number of parameters $P$, where a separate tree is grown for each parameter.\footnote{We want to stress that multi-parameter optimization is a known scaling problem of XGBoost and LightGBM for multiclass-classification and not restricted to our approach only.} To achieve a flexible approximation to the data, the order of the Bernstein-Polynomial $M$ needs to be reasonably high. As for the protein that shows the highest runtime, we use $P=M(10)+4$ parameters which requires the NFBoost models to estimate 7 times as many parameters compared to GBMLSS models. Another reason might be that NFBoost models rely on a set of TensorFlow functions that require many intermediate results to be shuffled back and forth between tensors and arrays. This can induce a significant latency, which in turn increases runtime. Furthermore, while all other models use exact gradients and hessians, NFBoost relies on automatic differentiation as an approximation. Although this greatly enhances the variety of loss functions available for model training, it might also have a negative effect on runtime. From experiments we have conducted so far it appears that the automatic derivation of the hessian poses a significant computational bottleneck. To circumvent approximating the second-order derivative, we also trained models in which the hessian was set to 1. However, this generally resulted in lower accuracy, which is consistent with the results reported by \citet{Sigrist.2021}.

%\begin{sidewaysfigure}[h!]
%	\centering
%	\includegraphics[width=1.0\linewidth]{UCI_CRPS_Normalized.pdf}
%	\caption{CRPS scores for the UCI datasets. For each dataset, all CRPS scores are normalized with respect to the median score of the best performing model.}
%	\label{fig:uci_crps_scores}
%\end{sidewaysfigure}

%%%%%%%%%%%%%%%%%%%%%%%%%%%%%%%%%%%%%%%%%%%%%%%%%%%%%%%%%%%%%%%%%%%%%%%%%%%%%%%%%
\section{Conclusion, Limitations and Future Research} \label{sec:conclusion}

\begin{quote} 
	\it{'Practitioners expect forecasting to reduce future uncertainty by providing accurate predictions like those in hard sciences. However, this is a great misconception. A major purpose of forecasting is not to reduce uncertainty but reveal its full extent and implications by estimating it as precisely as possible. [$\ldots$] The challenge for the forecasting field is how to persuade practitioners of the reality that all forecasts are uncertain and that this uncertainty cannot be ignored, as doing so could lead to catastrophic consequences.'} \citep{Makridakis.2021}
\end{quote}

\vspace{1em}

\noindent The language of statistics and machine learning is of probabilistic nature. Instead of a single point forecast only, distributional modelling provides a range of outcomes and the probability of each of those occurring. Consequently, any model that falls short of providing quantification of the uncertainty attached to its outcome is likely to yield an incomplete and potentially misleading picture. In an effort to obtain probabilistic forecasts from tree-based models, this paper presents a unified distributional gradient boosting framework for regression tasks that allows one to either model all conditional moments of a parametric distribution, or to approximate the conditional cumulative distribution function via Normalizing Flows. Based on a simulation study and real-world data examples, we have demonstrated our framework to be competitive to existing approaches and that Normalizing Flow based modelling of the conditional cumulative distribution function is superior in cases where the data cannot be described well using parametric distributions. Furthermore, our results suggest that the effectiveness of our Normalizing-Flow based tree-models tends to increase with data size.

Despite its flexibility, we acknowledge some limitations of our framework that require additional research. Even though distributional modelling relaxes the assumption of observations being identically distributed, tree-based models are not yet able to explicitly incorporate dependencies between observations, e.g., time, longitudinal or space. While features that represent, e.g., time, can be manually engineered, most tree-based models in their current implementation, however, are not able to infer these dependencies in the data themselves without appropriate changes of the estimation process. This contrasts with, for example, deep RNN models that are designed to model sequential data more naturally. Hence, a future extension of our approach would be to directly model longitudinal or temporal dependencies as part of the training process, as discussed in, e.g., \citet{Sela.2012, Hajjem.2011}. Another interesting extension of distributional modelling, as proposed by \citet{OMalley.2021, Klein.2020, Marra.2017, Klein.2016}, is to extend the univariate case to a multiple response setting, with several responses of interest that are potentially interdependent. Also, since our framework relies on multi-parameter optimization, where a separate tree is grown for each parameter, estimating many parameters for a large dataset can become computationally expensive, especially for NFBoost, where $M$ Bernstein-Polynomials in addition to 4 scale and shift parameters need to be estimated. For the time being, we leave a more runtime efficient version of our framework to future implementation and research. Finally, we consider the extension of our framework to allow for other Normalizing Flow types, suitable for count or ordinal data as proposed by \cite{Kook.2020}, as an interesting enhancement.

%%%%%%%%%%%%%%%%%%%%%%%%%%%%%%%%%%%%%%%%%%%%%%%%%%%%%%%%%%%%%%%%%%%%%%%%%%%%%%%%
\bibliography{literature}

\begin{thebibliography}{}

\bibitem[Abadi et~al., 2015]{Abadi.2015}
Abadi, M., Agarwal, A., Barham, P., Brevdo, E., Chen, Z., Citro, C., Corrado,
  G.~S., Davis, A., Dean, J., Devin, M., Ghemawat, S., Goodfellow, I., Harp,
  A., Irving, G., Isard, M., Jia, Y., Jozefowicz, R., Kaiser, L., Kudlur, M.,
  Levenberg, J., Man{\'e}, D., Monga, R., Moore, S., Murray, D., Olah, C.,
  Schuster, M., Shlens, J., Steiner, B., Sutskever, I., Talwar, K., Tucker, P.,
  Vanhoucke, V., Vasudevan, V., Vi{\'e}gas, F., Vinyals, O., Warden, P.,
  Wattenberg, M., Wicke, M., Yu, Y., and Zheng, X. (2015).
\newblock {TensorFlow: Large-Scale Machine Learning on Heterogeneous Systems}.

\bibitem[Akiba et~al., 2019]{Akiba.2019}
Akiba, T., Sano, S., Yanase, T., Ohta, T., and Koyama, M. (2019).
\newblock {Optuna: A Next-generation Hyperparameter Optimization Framework}.
\newblock In Teredesai, A., editor, {\em {Proceedings of the 25th ACM SIGKDD
  International Conference on Knowledge Discovery {\&} Data Mining}}, ACM
  Digital Library, pages 2623--2631, New York,NY,United States. {Association
  for Computing Machinery}.

\bibitem[Arpogaus, 2022]{Arpogaus.2022}
Arpogaus, M. (2022).
\newblock {Short-Term Probabilistic Load Forecasting using Conditioned
  Bernstein-Polynomial Normalizing Flows}.

\bibitem[Arpogaus et~al., 2021]{Arpogaus.2021}
Arpogaus, M., Voss, M., Sick, B., Nigge-Uricher, M., and D{\"u}rr, O. (2021).
\newblock {Probabilistic Short-Term Low-Voltage Load Forecasting using
  Bernstein-Polynomial Normalizing Flows}.
\newblock In {\em {ICML 2021, Workshop Tackling Climate Change with Machine
  Learning, June 26, 2021, virtual}}.

\bibitem[Athey et~al., 2019]{Athey.2019}
Athey, S., Tibshirani, J., and Wager, S. (2019).
\newblock {Generalized random forests}.
\newblock {\em {The Annals of Statistics}}, 47(2):1148--1178.

\bibitem[Baumann et~al., 2021]{Baumann.2021}
Baumann, P. F.~M., Hothorn, T., and R{\"u}gamer, D. (2021).
\newblock {Deep Conditional Transformation Models}.
\newblock In Oliver, N., {Perez Cruz}, F., Kramer, S., Read, J., and Lozano,
  J.~A., editors, {\em {Research Track}}, volume 12977 of {\em Lecture Notes in
  Artificial Intelligence}, pages 3--18. Springer, Cham.

\bibitem[Chen and Guestrin, 2016]{Chen.2016}
Chen, T. and Guestrin, C. (2016).
\newblock {XGBoost: A Scalable Tree Boosting System}.
\newblock In {\em {Proceedings of the 22nd ACM SIGKDD International Conference
  on Knowledge Discovery and Data Mining}}, KDD '16, pages 785--794, New York,
  NY, USA. {Association for Computing Machinery}.

\bibitem[Chipman et~al., 2010]{Chipman.2010}
Chipman, H.~A., George, E.~I., and McCulloch, R.~E. (2010).
\newblock {BART: Bayesian additive regression trees}.
\newblock {\em {The Annals of Applied Statistics}}, 4(1):266--298.

\bibitem[Dua and Graff, 2017]{Dua.2017}
Dua, D. and Graff, C. (2017).
\newblock {UCI Machine Learning Repository}.

\bibitem[Duan et~al., 2020]{Duan.2020}
Duan, T., Anand, A., Ding, D.~Y., Thai, K.~K., Basu, S., Ng, A.~Y., and
  Schuler, A. (2020).
\newblock {NGBoost: Natural Gradient Boosting for Probabilistic Prediction}.
\newblock In {\em {Proceedings of the 37th International Conference on Machine
  Learning, ICML 2020, 13-18 July 2020, Virtual Event}}, volume 119 of {\em
  Proceedings of Machine Learning Research}, pages 2690--2700. PMLR.

\bibitem[D{\"u}rr et~al., 2022]{Duerr.2022}
D{\"u}rr, O., H{\"o}rling, S., Dold, D., Kovylov, I., and Sick, B. (2022).
\newblock {Bernstein Flows for Flexible Posteriors in Variational Bayes}.
\newblock {\em {arXiv Pre-Print}}, pages 1--18.

\bibitem[Fahrmeir and Kneib, 2011]{Fahrmeir.2011}
Fahrmeir, L. and Kneib, T. (2011).
\newblock {\em {Bayesian smoothing and regression for longitudinal, spatial and
  event history data}}, volume~36 of {\em {Oxford statistical science series}}.
\newblock {Oxford University Press}, Oxford and New York.

\bibitem[Fahrmeir et~al., 2013]{Fahrmeir.2013}
Fahrmeir, L., Kneib, T., Lang, S., and Marx, B. (2013).
\newblock {\em {Regression: Models, methods and applications}}.
\newblock Springer, Berlin, 1 edition.

\bibitem[Farouki, 2012]{Farouki.2012}
Farouki, R.~T. (2012).
\newblock {The Bernstein polynomial basis: A centennial retrospective}.
\newblock {\em {Computer Aided Geometric Design}}, 29(6):379--419.

\bibitem[Friedman, 2020]{Friedman.2020}
Friedman, J.~H. (2020).
\newblock {Contrast trees and distribution boosting}.
\newblock {\em {Proceedings of the National Academy of Sciences}},
  117(35):21175--21184.

\bibitem[Gasthaus et~al., 2019]{Gasthaus.2019}
Gasthaus, J., Benidis, K., Wang, Y., Rangapuram, S.~S., Salinas, D., Flunkert,
  V., and Januschowski, T. (2019).
\newblock {Probabilistic Forecasting with Spline Quantile Function RNNs}.
\newblock In Chaudhuri, K. and Sugiyama, M., editors, {\em {Proceedings of the
  Twenty-Second International Conference on Artificial Intelligence and
  Statistics}}, volume~89 of {\em Proceedings of Machine Learning Research},
  pages 1901--1910. PMLR.

\bibitem[Giaquinto and Banerjee, 2020]{Giaquinto.2020}
Giaquinto, R. and Banerjee, A. (2020).
\newblock {Gradient Boosted Normalizing Flows}.
\newblock In {H. Larochelle}, {M. Ranzato}, {R. Hadsell}, {M. F. Balcan}, and
  {H. Lin}, editors, {\em {Advances in Neural Information Processing Systems}},
  volume~33, pages 22104--22117. {Curran Associates, Inc}.

\bibitem[Glorot and Bengio, 2010]{Glorot.2010}
Glorot, X. and Bengio, Y. (2010).
\newblock {Understanding the difficulty of training deep feedforward neural
  networks}.
\newblock In Teh, Y.~W. and Titterington, M., editors, {\em {Proceedings of the
  Thirteenth International Conference on Artificial Intelligence and
  Statistics}}, volume~9 of {\em Proceedings of Machine Learning Research},
  pages 249--256, Chia Laguna Resort, Sardinia, Italy. PMLR.

\bibitem[Gneiting and Raftery, 2007]{Gneiting.2007}
Gneiting, T. and Raftery, A.~E. (2007).
\newblock {Strictly proper scoring rules, prediction, and estimation}.
\newblock {\em {Journal of the American Statistical Association}},
  102(477):359--378.

\bibitem[Hajjem et~al., 2011]{Hajjem.2011}
Hajjem, A., Bellavance, F., and Larocque, D. (2011).
\newblock {Mixed effects regression trees for clustered data}.
\newblock {\em {Statistics {\&} Probability Letters}}, 81(4):451--459.

\bibitem[Hasson et~al., 2021]{Hasson.2021}
Hasson, H., Wang, B., Januschowski, T., and Gasthaus, J. (2021).
\newblock {Probabilistic Forecasting: A Level-Set Approach}.
\newblock In Beygelzimer, A., Dauphin, Y., Liang, P., and {Wortman Vaughan},
  J., editors, {\em {Advances in Neural Information Processing Systems}}.

\bibitem[He et~al., 2015]{He.2015}
He, K., Zhang, X., Ren, S., and Sun, J. (2015).
\newblock {Delving Deep into Rectifiers: Surpassing Human-Level Performance on
  ImageNet Classification}.
\newblock {\em {2015 IEEE International Conference on Computer Vision (ICCV)}},
  pages 1026--1034.

\bibitem[Hothorn, 2020]{Hothorn.2019}
Hothorn, T. (2020).
\newblock {Transformation Boosting Machines}.
\newblock {\em {Statistics and Computing}}, 30(1):141--152.

\bibitem[Hothorn et~al., 2014]{Hothorn.2014}
Hothorn, T., Kneib, T., and B{\"u}hlmann, P. (2014).
\newblock {Conditional transformation models}.
\newblock {\em {Journal of the Royal Statistical Society: Series B (Statistical
  Methodology)}}, 76(1):3--27.

\bibitem[Hothorn and Zeileis, 2021]{Hothorn.2021}
Hothorn, T. and Zeileis, A. (2021).
\newblock {Predictive Distribution Modeling Using Transformation Forests}.
\newblock {\em {Journal of Computational and Graphical Statistics}},
  30(4):1181--1196.

\bibitem[H{\"u}llermeier and Waegeman, 2021]{Hullermeier.2021}
H{\"u}llermeier, E. and Waegeman, W. (2021).
\newblock {Aleatoric and epistemic uncertainty in machine learning: an
  introduction to concepts and methods}.
\newblock {\em {Machine Learning}}, 110(3):457--506.

\bibitem[Januschowski et~al., 2020]{Januschowski.2020}
Januschowski, T., Gasthaus, J., Wang, Y., Salinas, D., Flunkert, V.,
  Bohlke-Schneider, M., and Callot, L. (2020).
\newblock {Criteria for classifying forecasting methods}.
\newblock {\em {International Journal of Forecasting}}, 36(1):167--177.

\bibitem[Januschowski and Kolassa, 2019]{Januschowski.2019}
Januschowski, T. and Kolassa, S. (2019).
\newblock {A classification of business forecasting problems}.
\newblock {\em {Foresight: The International Journal of Applied Forecasting}},
  52:36--43.

\bibitem[Januschowski et~al., 2021]{Januschowski.2021}
Januschowski, T., Wang, Y., Torkkola, K., Erkkil{\"a}, T., Hasson, H., and
  Gasthaus, J. (2021).
\newblock {Forecasting with trees}.
\newblock {\em {International Journal of Forecasting}}.

\bibitem[Ke et~al., 2017]{Ke.2017}
Ke, G., Meng, Q., Finley, T., Wang, T., Chen, W., Ma, W., Ye, Q., and Liu,
  T.-Y. (2017).
\newblock {LightGBM: A Highly Efficient Gradient Boosting Decision Tree}.
\newblock In {\em {Proceedings of the 31st International Conference on Neural
  Information Processing Systems}}, NIPS'17, pages 3149--3157, Red Hook, NY,
  USA. {Curran Associates Inc}.

\bibitem[Klein et~al., 2022]{Klein.2020}
Klein, N., Hothorn, T., Barbanti, L., and Kneib, T. (2022).
\newblock {Multivariate conditional transformation models}.
\newblock {\em {Scandinavian Journal of Statistics}}, 49(1):116--142.

\bibitem[Klein and Kneib, 2016]{Klein.2016}
Klein, N. and Kneib, T. (2016).
\newblock {Simultaneous inference in structured additive conditional copula
  regression models: a unifying Bayesian approach}.
\newblock {\em {Statistics and Computing}}, 26(4):841--860.

\bibitem[Klein et~al., 2015a]{Klein.2015c}
Klein, N., Kneib, T., and Lang, S. (2015a).
\newblock {Bayesian Generalized Additive Models for Location, Scale, and Shape
  for Zero-Inflated and Overdispersed Count Data}.
\newblock {\em {Journal of the American Statistical Association}},
  110(509):405--419.

\bibitem[Klein et~al., 2015b]{Klein.2015b}
Klein, N., Kneib, T., Lang, S., and Sohn, A. (2015b).
\newblock {Bayesian structured additive distributional regression with an
  application to regional income inequality in Germany}.
\newblock {\em {The Annals of Applied Statistics}}, 9(2):1024--1052.

\bibitem[Kobyzev et~al., 2020]{Kobyzev.2020}
Kobyzev, I., Prince, S., and Brubaker, M. (2020).
\newblock {Normalizing Flows: An Introduction and Review of Current Methods}.
\newblock {\em {IEEE Transactions on Pattern Analysis and Machine
  Intelligence}}, PP:1.

\bibitem[Koenker and Bassett, 1978]{Koenker.1978}
Koenker, R. and Bassett, G. (1978).
\newblock {Regression Quantiles}.
\newblock {\em {Econometrica}}, 46(1):33--50.

\bibitem[Kook et~al., 2021]{Kook.2020}
Kook, L., Herzog, L., Hothorn, T., D{\"u}rr, O., and Sick, B. (2021).
\newblock {Deep and interpretable regression models for ordinal outcomes}.
\newblock {\em {arXiv Pre-Print}}, pages 1--41.

\bibitem[Kurle et~al., 2020]{Kurle.2020}
Kurle, R., Rangapuram, S.~S., de~B{\'e}zenac, E., G{\"u}nnemann, S., and
  Gasthaus, J. (2020).
\newblock {Deep Rao-Blackwellised Particle Filters for Time Series
  Forecasting}.
\newblock In {H. Larochelle}, {M. Ranzato}, {R. Hadsell}, {M. F. Balcan}, and
  {H. Lin}, editors, {\em {Advances in Neural Information Processing Systems}},
  volume~33, pages 15371--15382. {Curran Associates, Inc}.

\bibitem[Lundberg et~al., 2020]{Lundberg.2020}
Lundberg, S.~M., Erion, G., Chen, H., DeGrave, A., Prutkin, J.~M., Nair, B.,
  Katz, R., Himmelfarb, J., Bansal, N., and Lee, S.-I. (2020).
\newblock {From local explanations to global understanding with explainable AI
  for trees}.
\newblock {\em {Nature Machine Intelligence}}, 2(1):2522--5839.

\bibitem[Lundberg and Lee, 2017]{Lundberg.2017}
Lundberg, S.~M. and Lee, S.-I. (2017).
\newblock {A Unified Approach to Interpreting Model Predictions}.
\newblock In {I. Guyon}, {U. V. Luxburg}, {S. Bengio}, {H. Wallach}, {R.
  Fergus}, {S. Vishwanathan}, and {R. Garnett}, editors, {\em {Advances in
  Neural Information Processing Systems 30}}, pages 4765--4774. {Curran
  Associates, Inc}.

\bibitem[Makridakis et~al., 2021a]{Makridakis.2021a}
Makridakis, S., Spiliotis, E., and Assimakopoulos, V. (2021a).
\newblock {The M5 competition: Background, organization, and implementation}.
\newblock {\em {International Journal of Forecasting}}.

\bibitem[Makridakis et~al., 2021b]{Makridakis.2021}
Makridakis, S., Spiliotis, E., Assimakopoulos, V., Chen, Z., Gaba, A., Tsetlin,
  I., and Winkler, R.~L. (2021b).
\newblock {The M5 uncertainty competition: Results, findings and conclusions}.
\newblock {\em {International Journal of Forecasting}}.

\bibitem[Marra and Radice, 2017]{Marra.2017}
Marra, G. and Radice, R. (2017).
\newblock {Bivariate copula additive models for location, scale and shape}.
\newblock {\em {Computational Statistics {\&} Data Analysis}}, 112:99--113.

\bibitem[M{\"a}rz, 2019]{Maerz.2019}
M{\"a}rz, A. (2019).
\newblock {XGBoostLSS - An extension of XGBoost to probabilistic forecasting}.
\newblock {\em {arXiv Pre-Print}}, pages 1--23.

\bibitem[M{\"a}rz, 2020]{Maerz.2020}
M{\"a}rz, A. (2020).
\newblock {CatBoostLSS - An extension of CatBoost to probabilistic
  forecasting}.
\newblock {\em {arXiv Pre-Print}}, pages 1--18.

\bibitem[Meinshausen, 2006]{Meinshausen.2006}
Meinshausen, N. (2006).
\newblock {Quantile Regression Forests}.
\newblock {\em {Journal of Machine Learning Research}}, 7(35):983--999.

\bibitem[O'Malley et~al., 2021]{OMalley.2021}
O'Malley, M., Sykulski, A.~M., Lumpkin, R., and Schuler, A. (2021).
\newblock {Multivariate Probabilistic Regression with Natural Gradient
  Boosting}.
\newblock {\em {arXiv Pre-Print}}, pages 1--19.

\bibitem[Papamakarios et~al., 2021]{Papamakarios.2021}
Papamakarios, G., Nalisnick, E.~T., Rezende, D.~J., Mohamed, S., and
  Lakshminarayanan, B. (2021).
\newblock {Normalizing Flows for Probabilistic Modeling and Inference}.
\newblock {\em {Journal of Machine Learning Research}}, 22:1--64.

\bibitem[Paszke et~al., 2019]{Paszke.2019}
Paszke, A., Gross, S., Massa, F., Lerer, A., Bradbury, J., Chanan, G., Killeen,
  T., Lin, Z., Gimelshein, N., Antiga, L., Desmaison, A., K{\"o}pf, A., Yang,
  E., DeVito, Z., Raison, M., Tejani, A., Chilamkurthy, S., Steiner, B., Fang,
  L., Bai, J., and Chintala, S. (2019).
\newblock {PyTorch: An Imperative Style, High-Performance Deep Learning
  Library}.
\newblock In {\em {Proceedings of the 33rd International Conference on Neural
  Information Processing Systems}}. {Curran Associates Inc}, Red Hook, NY, USA.

\bibitem[Pratola et~al., 2020]{Pratola.2020}
Pratola, M.~T., Chipman, H.~A., George, E.~I., and McCulloch, R.~E. (2020).
\newblock {Heteroscedastic BART via Multiplicative Regression Trees}.
\newblock {\em {Journal of Computational and Graphical Statistics}},
  29(2):405--417.

\bibitem[Rangapuram et~al., 2021]{Rangapuram.2021}
Rangapuram, S.~S., Werner, L.~D., Benidis, K., Mercado, P., Gasthaus, J., and
  Januschowski, T. (2021).
\newblock {End-to-End Learning of Coherent Probabilistic Forecasts for
  Hierarchical Time Series}.
\newblock In Meila, M. and Zhang, T., editors, {\em {Proceedings of the 38th
  International Conference on Machine Learning}}, volume 139 of {\em
  Proceedings of Machine Learning Research}, pages 8832--8843. PMLR.

\bibitem[Rasul et~al., 2021]{Rasul.2021}
Rasul, K., Sheikh, A.-S., Schuster, I., Bergmann, U.~M., and Vollgraf, R.
  (2021).
\newblock {Multivariate Probabilistic Time Series Forecasting via Conditioned
  Normalizing Flows}.
\newblock In {\em {International Conference on Learning Representations}}.

\bibitem[Rigby and Stasinopoulos, 2005]{Rigby.2005}
Rigby, R.~A. and Stasinopoulos, D.~M. (2005).
\newblock {Generalized additive models for location, scale and shape}.
\newblock {\em {Journal of the Royal Statistical Society: Series C (Applied
  Statistics)}}, 54(3):507--554.

\bibitem[R{\"u}gamer et~al., 2022]{Ruegamer.2022}
R{\"u}gamer, D., Baumann, P. F.~M., Kneib, T., and Hothorn, T. (2022).
\newblock {Probabilistic Time Series Forecasts with Autoregressive
  Transformation Models}.
\newblock {\em {arXiv Pre-Print}}, pages 1--16.

\bibitem[Salinas et~al., 2020]{Salinas.2020}
Salinas, D., Flunkert, V., Gasthaus, J., and Januschowski, T. (2020).
\newblock {DeepAR: Probabilistic forecasting with autoregressive recurrent
  networks}.
\newblock {\em {International Journal of Forecasting}}, 36(3):1181--1191.

\bibitem[Schlosser et~al., 2018]{Schlosser.2018}
Schlosser, L., Hothorn, T., Stauffer, R., and Zeileis, A. (2018).
\newblock {Distributional Regression Forests for Probabilistic Precipitation
  Forecasting in Complex Terrain}.
\newblock {\em {The Annals of Applied Statistics}}, 13(3):1564--1589.

\bibitem[Sela and Simonoff, 2012]{Sela.2012}
Sela, R.~J. and Simonoff, J.~S. (2012).
\newblock {RE-EM trees: a data mining approach for longitudinal and clustered
  data}.
\newblock {\em {Machine Learning}}, 86(2):169--207.

\bibitem[Sick et~al., 2021]{Sick.2021}
Sick, B., Hothorn, T., and D{\"u}rr, O. (2021).
\newblock {Deep transformation models: Tackling complex regression problems
  with neural network based transformation models}.
\newblock In {\em {2020 25th International Conference on Pattern Recognition
  (ICPR)}}, pages 2476--2481.

\bibitem[Sigrist, 2021]{Sigrist.2021}
Sigrist, F. (2021).
\newblock {Gradient and Newton boosting for classification and regression}.
\newblock {\em {Expert Systems with Applications}}, 167:114080.

\bibitem[Sprangers et~al., 2021]{Sprangers.2021}
Sprangers, O., Schelter, S., and de~Rijke, M. (2021).
\newblock {Probabilistic Gradient Boosting Machines for Large-Scale
  Probabilistic Regression}.
\newblock In {\em {Proceedings of the 27th ACM SIGKDD Conference on Knowledge
  Discovery {\&} Data Mining}}, KDD '21, pages 1510--1520, New York, NY, USA.
  {Association for Computing Machinery}.

\bibitem[Stasinopoulos et~al., 2017]{Stasinopoulos.2017}
Stasinopoulos, M.~D., Rigby, R.~A., Heller, G.~Z., Voudouris, V., and
  de~Bastiani, F. (2017).
\newblock {\em {Flexible Regression and Smoothing: Using GAMLSS in R}}.
\newblock {Chapman {\&} Hall / CRC The R Series}. {CRC Press}, London.

\bibitem[Ziel, 2021]{Ziel.2021}
Ziel, F. (2021).
\newblock {M5 competition uncertainty: Overdispersion, distributional
  forecasting, GAMLSS, and beyond}.
\newblock {\em {International Journal of Forecasting}}.

\end{thebibliography}

%%%%%%%%%%%%%%%%%%%%%%%%%%%%%%%%%%%%%%%%%%%%%%%%%%%%%%%%%%%%%%%%%%%%%%%%%%%%%%%%

\end{document}